\documentclass{article}

    \PassOptionsToPackage{numbers, compress}{natbib}

 \usepackage[main, final]{neurips_2026}

\usepackage[utf8]{inputenc} 
\usepackage[T1]{fontenc}    
\usepackage{hyperref}       
\usepackage{url}            
\usepackage{booktabs}       
\usepackage{amsfonts}       
\usepackage{nicefrac}       
\usepackage{microtype}      
\usepackage{xcolor}         

\usepackage{graphicx}
\usepackage{amsmath}
\usepackage{amssymb}
\usepackage{mathtools}
\usepackage{amsthm}
\usepackage{algorithm}
\usepackage{algorithmic}
\usepackage[table]{xcolor}
\usepackage{multirow}
\usepackage{arydshln}
\usepackage{subcaption}
\usepackage{tabularx}
\usepackage{tcolorbox}
\usepackage{enumitem}
\usepackage{titletoc}
\usepackage{cleveref}
\usepackage{wrapfig}

\newcommand{\method}{\textsc{SkillRL}}
\newcommand{\skillbank}{\textsc{SkillBank}}

\title{\method{}: Evolving Agents via Recursive Skill-Augmented Reinforcement Learning}

\author{Peng Xia$^{1}$\thanks{Equal Contribution.}, Jianwen Chen$^{1*}$, Hanyang Wang$^{12*}$, Jiaqi Liu$^{1}$, Kaide Zeng$^{1}$, \\ \textbf{Yu Wang$^{3}$, Siwei Han$^{1}$, Yiyang Zhou$^{1}$, Xujiang Zhao$^{4}$, Haifeng Chen$^{4}$,}\\ \textbf{Zeyu Zheng$^{5}$, Cihang Xie$^{6}$, Huaxiu Yao$^{1}$} \\ $^1$UNC-Chapel Hill, $^2$University of Chicago, $^3$UC San Diego, \\ $^4$NEC Labs America, $^5$UC Berkeley, $^6$UC Santa Cruz \\ \texttt{\{pxia,huaxiu\}@cs.unc.edu}}

\begin{document}

\maketitle

\begin{abstract}
Large Language Model (LLM) agents have shown stunning results in complex tasks, yet they often operate in isolation, failing to learn from past experiences. Existing memory-based methods primarily store raw trajectories, which are often redundant and noise-heavy. This prevents agents from extracting high-level, reusable behavioral patterns that are essential for generalization. In this paper, we propose \method{}, a framework that bridges the gap between raw experience and policy improvement through automatic skill discovery and recursive evolution. Our approach introduces an experience-based distillation mechanism to build a hierarchical skill library \skillbank, an adaptive retrieval strategy for general and task-specific heuristics, and a recursive evolution mechanism that allows the skill library to co-evolve with the agent's policy during reinforcement learning. These innovations significantly reduce the token footprint while enhancing reasoning utility. Experimental results on ALFWorld, WebShop and seven search-augmented tasks demonstrate that \method{} achieves state-of-the-art performance, outperforming strong baselines over 15.3\% and maintaining robustness as task complexity increases. Code is available at this \href{https://github.com/aiming-lab/SkillRL}{https://github.com/aiming-lab/SkillRL}.
\end{abstract}

\section{Introduction}
\label{sec:intro}
Large language model (LLM) agents~\cite{yao2022react,shinn2023reflexion} have demonstrated remarkable capabilities across various sophisticated tasks~\cite{xia2026rrsi,huang2026envharness}, such as web navigation~\cite{geminigui,openaigui} and deep research~\cite{openaideepresearch,geminideepresearch,team2025tongyi}, by interacting with complex environments through natural language. Despite these advances, each task execution remains largely episodic. Current LLM agents operate in isolation, unable to learn from past successes or failures~\cite{zhang2025memevolve}, which significantly hinders their evolution. Consequently, a fundamental challenge remains: \emph{how can agents efficiently learn from experience and transfer that knowledge to other tasks?}

\begin{wrapfigure}{r}{0.6\textwidth}
    \centering
    \includegraphics[width=0.59\textwidth]{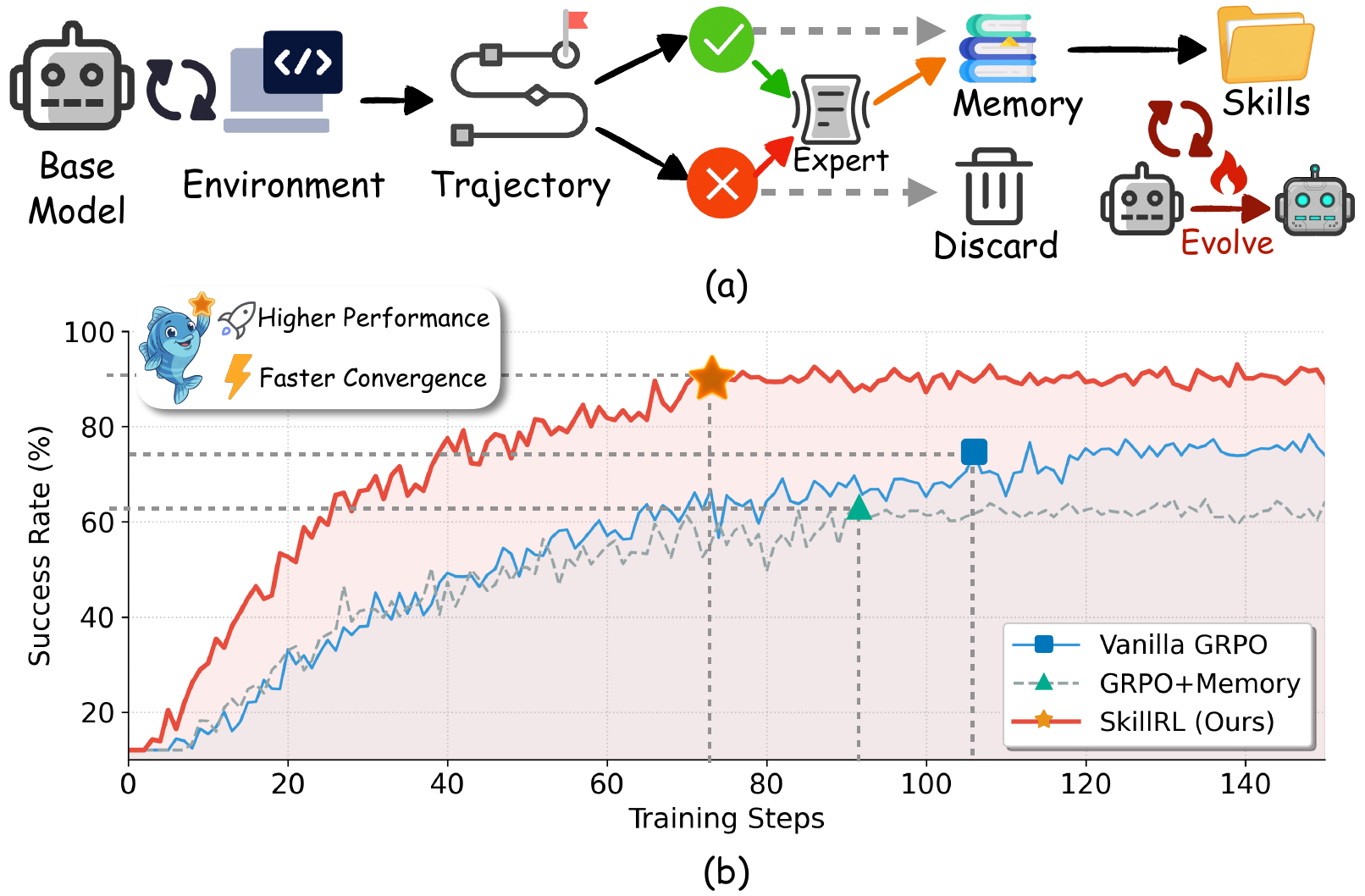}
    \caption{(a) Overview of the \method{} pipeline. Unlike previous methods (\textcolor{gray}{gray dashed lines}) that store raw trajectories and discard failures, \method{} employs an experience-based distillation mechanism to transform diverse experiences into structured skills. (b) Performance on ALFWorld validation set. \method{} achieves faster convergence and superior success rates compared to vanilla GRPO and memory-augmented RL.}
    \label{fig:intro}
    \vspace{-1em}
\end{wrapfigure}

The existing memory-based methods for LLM agents primarily involve saving raw trajectories directly into external databases during the sampling process to serve as references for similar future tasks~\cite{shinn2023reflexion,zhao2024expel}. While intuitive, these raw trajectories are often lengthy and contain significant redundancy and noise~\cite{chhikara2025mem0}, making it difficult for the model to extract critical information. Recent work has attempted to compress trajectories and update the memory bank via online training~\cite{zhang2025memevolve,zhang2026memrl}, improving memory efficiency. However, these methods merely mimic past solutions and they fail to distill core principles or adapt the agent’s internal policy to leverage memory for guided decision-making. As depicted in the dashed flow of Figure~\ref{fig:intro}(a), such approaches often struggle with the trade-off between information density and noise, leading to sub-optimal performance or even degradation as shown in Figure~\ref{fig:intro}(b).

We argue that these approaches miss a crucial insight: effective experience transfer requires \emph{abstraction}. Human experts do not memorize every action in every situation; instead, they develop \emph{skills}~\cite{authropic3}, compact and reusable strategies that capture the essence of how to accomplish specific subtasks. Inspired by this observation, we propose \method{}, a framework that bridges the gap between raw experience and efficient policy improvement through automatic skill discovery and recursive skill evolution.

\method{} first introduces an experience-based skill distillation mechanism, which gathers diverse trajectories from environment rollouts and applies differential processing: successful episodes are preserved as demonstrations, while failed ones are synthesized into concise failure lessons to mitigate context noise. Secondly, we transform these experiences into a hierarchical skill library \skillbank, differentiating between \emph{general skills} for universal strategic guidance and \emph{task-specific skills} for task-level heuristics. This abstraction allows the agent to adaptively retrieve relevant skills during decision-making, significantly reducing the token footprint while enhancing reasoning utility. Lastly, \method{} incorporates a recursive skill evolution mechanism during reinforcement learning (RL), where the skill library is treated as a dynamic component rather than a static knowledge source. By analyzing failure modes after each validation epoch to generate new skills or refine existing ones, our approach ensures the skill library and the agent’s policy co-evolve, maintaining robustness as task complexity increases. As demonstrated in Figure~\ref{fig:intro}(b), \method{} achieves substantially faster convergence and higher asymptotic performance.

The primary contribution is \method{}, a framework that enables LLM agents to bridge the gap between raw experience and policy improvement through automatic skill discovery and recursive evolution. By distilling redundant trajectories into a hierarchical \skillbank, our method abstracts general and task-specific skills to guide decision-making efficiently. Furthermore, we introduce a recursive evolution mechanism that ensures the skill library and agent policy co-evolve during reinforcement learning. Empirical results on ALFWorld, WebShop, and seven search-augmented benchmarks demonstrate that \method{} achieves state-of-the-art performance with 15.3\% improvements, significantly outperforming current memory-based agent-tuning baselines in both task success and reasoning utility.

\section{Preliminaries}
\label{sec:formulation}

\textbf{LLM Agents.} We consider an agent operating in an interactive environment $\mathcal{E}$. At each timestep $t$, the agent observes a state $o_t \in \mathcal{O}$, selects an action $a_t \in \mathcal{A}$, and receives a reward $r_t$ and next observation $o_{t+1}$. A trajectory $\tau = (o_0, a_0, r_0, \ldots, o_T, a_T, r_T)$ captures one episode of interaction. Tasks are specified by natural language descriptions $d$.
An LLM-based agent parameterized by $\theta$ implements a policy $\pi_\theta(a_t | o_{\leq t}, d, c)$ where $c$ represents additional context (e.g., skills, demonstrations). Our goal is to learn a policy that maximizes expected return $\small \max_\theta \mathbb{E}_{\tau \sim \pi_\theta} \left[ \sum_{t=0}^{T} \gamma^t r_t \right]$
subject to context length constraints $|c| \leq L_{\max}$.

\textbf{Group Relative Policy Optimization (GRPO).} GRPO~\cite{shao2024deepseekmath} is a reinforcement learning method that avoids training a critic by using intra-group relative rewards to optimize the policy. For each query $x$, the model samples $G$ responses $\{y^{(1)}, \ldots, y^{(G)}\}$, which are scored to obtain rewards $\{R_1, \ldots, R_G\}$. GRPO computes normalized advantages and updates the policy with a PPO-style clipped objective \cite{schulman2017proximal}:
\begin{equation}
\footnotesize
    \mathcal{J}_{\text{GRPO}}(\theta) = \mathbb{E}_{x, \{y_i\}} \Bigg[ \frac{1}{G} \sum_{i=1}^{G} \min \Big( r_i A_i,
    \text{clip}(r_i, 1-\epsilon, 1+\epsilon) A_i \Big) - \beta D_{\text{KL}}(\pi_\theta \| \pi_{\text{ref}}) \Bigg],
\end{equation}
where \begin{small}$\small r_i = \frac{\pi_\theta(y_i | x)}{\pi_{\text{old}}(y_i | x)}$\end{small} is the importance ratio, \begin{small}$\small A_i = \frac{R_i - \text{mean}(\{R_j\}_{j=1}^G)}{\text{std}(\{R_j\}_{j=1}^G)}$\end{small} is the normalized advantage, $\epsilon$, $\beta$ are hyperparameters, and $\pi_{\text{old}}$ is the policy before the current update. 

\vspace{-0.5em}
\section{\method{}}
\label{sec:method}

In this section, as illustrated in \Cref{fig:overview}, we propose \method{}, a framework designed to bridge the gap between raw interaction experience and policy improvement through automatic skill discovery and recursive evolution. \method{} consists of three core components. First, we develop an experience-based skill distillation mechanism to transform redundant trajectories into concise, actionable knowledge. Second, we organize these distilled experiences into a hierarchical skill library $\mathcal{S}$, enabling efficient retrieval of general and task-specific expertise. Lastly, we introduce a recursive skill evolution mechanism that leverages RL to dynamically refine the skill library in tandem with the agent’s policy. We detail these components as follows:

\begin{figure}[t]
    \centering
    \includegraphics[width=0.94\textwidth]{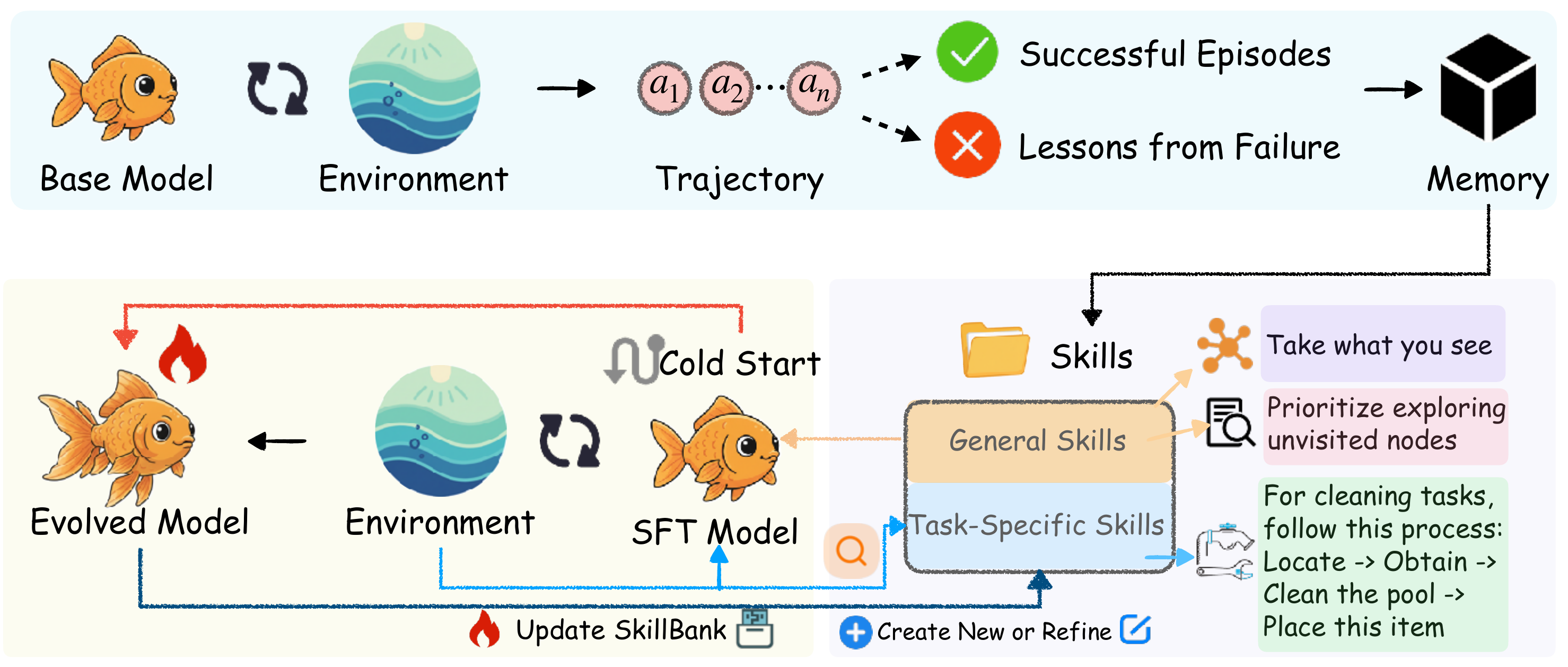}
    \caption{Overview of the \method{} framework. We collect trajectories using a base model, distill them into a hierarchical skill library, perform cold-start SFT to enable skill utilization, and then conduct RL training with dynamic skill evolution based on validation failures.}
    \label{fig:overview}
    \vspace{-1.5em}
\end{figure}

\vspace{-0.5em}
\subsection{Experience-based Skill Distillation}

Raw trajectories $\tau$ collected from environment interactions are verbose, containing exploratory actions, backtracking, and redundant steps that obscure the critical decisions leading to success or failure. To transform these experiences into actionable knowledge, we employ a teacher model $\mathcal{M}_T$ to distill trajectories into compact, reusable skills.

Specifically, we first deploy a base LLM agent $\pi_{\text{base}}$ in the target environment $\mathcal{E}$ to collect diverse trajectories. Unlike prior approaches that retain only successful episodes, we deliberately preserve both successful trajectories $\mathcal{T}^+ = \{\tau_i : r(\tau_i) = 1\}$ and failed trajectories $\mathcal{T}^- = \{\tau_i : r(\tau_i) = 0\}$, where $r(\tau)$ denotes the binary task success indicator. Failed trajectories reveal failure modes and boundary conditions, i.e., information difficult to infer from successes alone.

We apply differential processing based on trajectory outcomes. For \emph{successful trajectories} $\tau^+ \in \mathcal{T}^+$, we extract the strategic patterns that led to task completion:
\begin{equation} \small
    s^+ = \mathcal{M}_T(\tau^+, d).
\end{equation}
The teacher model identifies critical decision points, the reasoning behind correct actions, and generalizable patterns that transfer beyond the specific task instance.

For \emph{failed trajectories} $\tau^- \in \mathcal{T}^-$, direct inclusion in context is infeasible due to their length and noise. Instead, we synthesize concise failure lessons:
\begin{equation}
    s^- = \mathcal{M}_T(\tau^-, d).
\end{equation}
The analysis identifies: (1) the point of failure, (2) the flawed reasoning or action, (3) what should have been done, and (4) general principles to prevent similar failures. This transforms verbose failed episodes into counterfactuals.

\subsection{Hierarchical Skill Library (\skillbank) Construction}
Following the design principles of Agent Skills~\cite{authropic3}, we organize the distilled knowledge into a hierarchical skill library \skillbank{} that enables efficient retrieval of relevant expertise during decision-making.

\paragraph{Skill Organization.}
We structure \skillbank{} into two levels:
1) \emph{General Skills} $\mathcal{S}_g$ capture universal strategic principles applicable across all task types within an environment. These typically include exploration strategies (e.g., systematic search patterns, prioritizing unvisited locations), state management principles (e.g., verifying preconditions before actions), and goal-tracking heuristics (e.g., maintaining progress counters, terminating only upon verified completion). General skills provide foundational guidance that transfers across different task categories.
2) \emph{Task-Specific Skills} $\mathcal{S}_k$ encode specialized knowledge for task category $k$. These capture domain-specific action sequences, task-particular preconditions and constraints, common failure modes unique to the task type, and optimized procedures that exploit task structure. By organizing trajectories by task type during collection, we enable extraction of fine-grained, category-specific strategies that complement the broader general skills.

The complete skill library \skillbank\ is $\mathcal{S}_g \cup \bigcup_{k=1}^{K} \mathcal{S}_k$.
Each skill $s \in \skillbank{}$ is structured with: a concise name (e.g., systematic exploration), a principle describing the strategy, and when\_to\_apply conditions specifying applicability. This format enables efficient retrieval while providing clear guidance for application.

\paragraph{Skill Retrieval.}
At inference, given a task description $d$, the agent retrieves relevant skills to augment its context. General skills $\mathcal{S}_g$ are always included as foundational guidance. Task-specific skills are retrieved via semantic similarity:
\begin{equation}
    \mathcal{S}_{\text{ret}} = \text{TopK}\left(\{s \in \mathcal{S}_k : \text{sim}(e_d, e_s) > \delta\}, K\right),
\end{equation}
where $e_d, e_s$ are embeddings of the task description and skill respectively, $\delta$ is a similarity threshold, and $K$ controls the number of retrieved skills. The policy then conditions on the retrieved skills:
\begin{equation}
    a_t \sim \pi_\theta(a_t | o_{\leq t}, d, \mathcal{S}_g, \mathcal{S}_{\text{ret}}).
\end{equation}
Notably, skill distillation achieves 10--20$\times$ token compression compared to raw trajectories while enhancing rather than degrading the utility of the original experience. This compression allows the agent to leverage rich experiential knowledge within limited context windows.

\subsection{Recursive Skill Evolution}
A static skill library cannot anticipate all scenarios the agent will encounter. As the policy improves and explores new state regions, it faces situations where existing skills provide insufficient guidance. We introduce recursive skill evolution during reinforcement learning to address this limitation, enabling the skill library and agent policy to co-evolve.

\noindent \textbf{Cold-Start Initialization.}
Before RL training, we address a critical challenge: the base agent has not learned how to effectively utilize skills. Simply providing skills to an unchanged model yields limited benefit~\cite{guo2025deepseek}. We therefore perform a cold-start supervised fine-tuning (SFT) stage~\cite{ouyang2022training}, where the teacher model $\mathcal{M}_T$ generates $N$ skill-augmented reasoning traces $\mathcal{D}_{\text{SFT}} = \{(d_i, \mathcal{S}_i, \tau_i^*)\}_{i=1}^N$ demonstrating how to retrieve, interpret, and apply skills during decision-making. The base model is then fine-tuned on these demonstrations:
\begin{equation}
    \theta_{\text{sft}} = \arg\min_\theta \mathcal{L}_{\text{CE}}(\mathcal{D}_{\text{SFT}}; \theta),
\end{equation}
where $\mathcal{L}_{\text{CE}}$ denotes the cross-entropy loss. The resulting model $\pi_{\theta_{\text{sft}}}$ serves as both the starting point for RL training and the reference policy $\pi_{\text{ref}}$ for KL regularization.

\noindent \textbf{Recursive Skill Evolution.}
A static skill library cannot anticipate all scenarios the agent will encounter. As the policy improves and explores new state regions, it faces situations where existing skills provide insufficient guidance. We introduce recursive skill evolution to address this limitation. The process begins with an initial skill library containing baseline task-action principles.

After each validation epoch, we monitor the success rate $Acc(C)$ for each task category $C$. To ensure targeted growth, the evolution is triggered only for categories where $Acc(C) < \delta$. We then collect failed trajectories $\mathcal{T}_{\text{val}}^- = \{\tau_j : r(\tau_j) = 0\}_{j=1}^{M}$ using a diversity-aware stratified sampling strategy: trajectories are grouped by category, prioritized by the severity of failure (negative rewards), and selected via round-robin sampling to maintain categorical entropy. Then we will analyze these samples to identify gaps:
\begin{equation}
    \mathcal{S}_{\text{new}} = \mathcal{M}_T(\mathcal{T}_{\text{val}}^-, \skillbank{}).
\end{equation}
The teacher model is prompted to: (1) identify failure patterns not addressed by current skills, (2) propose new skills to cover these gaps, and (3) suggest refinements to existing skills that proved ineffective. The library is then updated: $\skillbank{} \leftarrow \skillbank{} \cup \mathcal{S}_{\text{new}}$.

This creates a virtuous cycle: as the agent improves, it encounters new challenges, which drive skill library expansion, which enables further improvement. 

\noindent \textbf{RL-based Policy Optimization.}
We optimize the skill-augmented policy using GRPO. For each task with description $d$, the agent first retrieves relevant skills and then samples $G$ complete trajectories $\{\tau^{(1)}, \ldots, \tau^{(G)}\}$ from the current policy $\pi_\theta$. Each trajectory $\tau^{(i)}$ receives a binary reward $R_i = r(\tau^{(i)}) \in \{0, 1\}$ indicating task successfulness. The normalized advantage for each trajectory is computed as:
\begin{equation} \footnotesize
    A_i = \frac{R_i - \text{mean}(\{R_j\}_{j=1}^G)}{\text{std}(\{R_j\}_{j=1}^G)}.
\end{equation}

The policy is updated according to:
\begin{equation}
\footnotesize
    \mathcal{J}(\theta) = \mathbb{E}_{d, \{\tau^{(i)}\}} \Bigg[ \frac{1}{G} \sum_{i=1}^{G} \min \Big( \rho_i A_i,
    \text{clip}(\rho_i, 1-\epsilon, 1+\epsilon) A_i \Big) - \beta D_{\text{KL}}(\pi_\theta \| \pi_{\text{ref}}) \Bigg],
\end{equation}
where $\rho_i = \frac{\pi_\theta(\tau^{(i)} | d, \mathcal{S}_g, \mathcal{S}_{\text{ret}})}{\pi_{\text{old}}(\tau^{(i)} | d, \mathcal{S}_g, \mathcal{S}_{\text{ret}})}$ is the importance ratio computed over the skill-augmented context. The KL penalty anchored to $\pi_{\text{ref}} = \pi_{\theta_{\text{sft}}}$ ensures that RL optimization preserves the learned skill utilization capabilities while improving task performance.
The complete training procedure is summarized in Algorithm~\ref{alg:skillrl}.

\section{Experiments}
\label{sec:experiments}

\begin{table}[ht]
\centering
\footnotesize
\caption{Performance on ALFWorld and WebShop. For ALFWorld, we report the average success rate (\%) for each subtask as well as the overall result. For WebShop, we report both the average score and the average success rate (\%). $^*$ denotes the results replicated from~\citep{feng2025group}. The best results and second best results are highlighted in \colorbox{red!25}{red} and \colorbox{blue!15}{blue}, respectively.}
\label{tab:performance}
\begin{tabular}{lccccccc|cc}
\toprule
\multirow{2}{*}{\textbf{Method}} & \multicolumn{7}{c|}{\textbf{ALFWorld}} & \multicolumn{2}{c}{\textbf{WebShop}} \\
 & Pick & Look & Clean & Heat & Cool & Pick2 & All & Score & Succ. \\
\midrule
\rowcolor{gray!15} \multicolumn{10}{l}{\textit{Closed-source LLMs}} \\
GPT-4o & 75.3 & 60.8 & 31.2 & 56.7 & 21.6 & 49.8 & 48.0 & 31.8 & 23.7 \\
 Gemini-2.5-Pro & 92.8 & 63.3 & 62.1 & 69.0 & 26.6 & 58.7 & 60.3 & 42.5 & 35.9 \\
\midrule
\textit{Qwen2.5-7B-Instruct} & & & & & & & & & \\
 Qwen2.5 & 33.4 & 21.6 & 19.3 & 6.90 & 2.80 & 3.20 & 14.8 & 26.4 & 7.80 \\
 \rowcolor{gray!15} \multicolumn{10}{l}{\textit{Prompt-based Agentic or Memory-based Methods}} \\
 ReAct$^*$ & 48.5 & 35.4 & 34.3 & 13.2 & 18.2 & 17.6 & 31.2 & 46.2 & 19.5 \\
 Reflexion$^*$ & 62.0 & 41.6 & 44.9 & 30.9 & 36.3 & 23.8 & 42.7 & 58.1 & 28.8 \\
 Mem0 & 54.0 & 55.0 & 26.9 & 36.4 & 20.8 & 7.69 & 33.6 & 23.9 & 2.00 \\ 
 ExpeL & 21.0 & 67.0 & 55.0 & 52.0 & 71.0 & 6.00 & 46.3 & 30.9 & 11.2 \\ 
 MemP & 54.3 & 38.5 & 48.1 & 56.2 & 32.0 & 16.7 & 41.4 & 25.3 & 6.40  \\
 SimpleMem & 64.5 & 33.3 & 20.0 & 12.5 & 33.3 & 3.84 & 29.7 & 33.2 & 8.59 \\
 \hdashline
 \rowcolor{gray!15} \multicolumn{10}{l}{\textit{RL-based Methods}} \\
 RLOO$^*$ & 87.6 & \cellcolor{red!25}{78.2} & 87.3 & \cellcolor{blue!15}{81.3} & 71.9 & 48.9 & 75.5 & \cellcolor{blue!15}{80.3} & 65.7 \\
 GRPO$^*$ & \cellcolor{blue!15}{90.8} & 66.1 & \cellcolor{blue!15}{89.3} & 74.7 & \cellcolor{blue!15}{72.5} & \cellcolor{blue!15}{64.7} & \cellcolor{blue!15}{77.6} & 79.3 & \cellcolor{blue!15}{66.1} \\
 \hdashline
  \rowcolor{gray!15} \multicolumn{10}{l}{\textit{Memory-Augmented RL-based Methods}} \\
  MemRL & 62.8 & 38.5 & 22.2 & 12.5 & 8.00 & 0.00 & 21.4 & 29.5 & 9.20 \\
 EvolveR & 64.9 & 33.3 & 46.4 & 13.3 & 33.3 & 33.3 & 43.8 & 42.5 & 17.6 \\
 Mem0+GRPO & 78.1 & 54.8 & 56.1 & 31.0 & 65.0 & 26.9 & 54.7 & 58.1 & 37.5 \\
 SimpleMem+GRPO & 89.5 & 36.3 & 60.0 &50.0& 64.9 & 26.3 & 62.5 & 67.8& 46.9 \\
 \method{} & \cellcolor{red!25}{97.9} & \cellcolor{blue!15}{71.4} & \cellcolor{red!25}{90.0} & \cellcolor{red!25}{90.0} & \cellcolor{red!25}{95.5} & \cellcolor{red!25}{87.5} & \cellcolor{red!25}{89.9} & \cellcolor{red!25}{85.2} & \cellcolor{red!25}{72.7} \\
\bottomrule
\end{tabular}
\end{table}

\begin{table*}[htbp]
\centering
\caption{Performance on search-augmented QA tasks. \method{} is trained on NQ and HotpotQA. $^\dagger$ and $^\star$ indicate in-domain and out-of-domain datasets, respectively. $^*$ denotes the results replicated from~\citep{sun2025zerosearch}.}
\label{tab:performance_search}
\footnotesize
\setlength{\tabcolsep}{4.5pt} 
\begin{tabular}{l ccc | cccc | c}
\toprule
\multirow{2}{*}{\textbf{Method}} & \multicolumn{3}{c|}{\textbf{Single-Hop QA}} & \multicolumn{4}{c|}{\textbf{Multi-Hop QA}} & \multirow{2}{*}{\textbf{Avg.}} \\
& NQ$^\dagger$ & TriviaQA$^\star$ & PopQA$^\star$ & HotpotQA$^\dagger$ & 2Wiki$^\star$ & MuSiQue$^\star$ & Bamboogle$^\star$ & \\
\midrule
\multicolumn{9}{l}{\textit{Qwen2.5-7B-Instruct}} \\
Qwen2.5$^*$ & 11.6 & 35.6 & 1.20 & 16.4 & 22.2 & 4.80 & 14.4 & 15.2 \\
CoT$^*$ & 12.8 & 35.6 & 3.80 & 16.2 & 22.6 & 6.60 & 24.0 & 17.4 \\
RAG$^*$ & 27.4 & 58.2 & 17.8 & 25.8 & 23.2 & 9.40 & 16.8 & 25.5 \\
Search-o1$^*$ & 19.4 & 40.6 & 11.4 & 17.0 & 27.0 & 8.60 & 30.4 & 22.1 \\ 
R1-Instruct & 21.0 & 44.9 & 17.1 & 20.8 & 27.5 & 6.00 & 19.2 & 22.4 \\
Search-R1 & 39.3 & 61.0 & 39.7 & 37.0 & 40.1 & 14.6 & 36.8 & 38.5 \\
ZeroSearch & \colorbox{blue!15}{43.6} & 61.8 & 51.5 & 34.6 & 35.2 & 18.4 & 27.8 & 39.1 \\
StepSearch & - & - & - & \colorbox{blue!15}{38.6} & 36.6 & \colorbox{red!25}{22.6} & 40.0 & - \\
EvolveR & 43.5 & \colorbox{red!25}{63.4} & \colorbox{blue!15}{44.6} & 38.2 & \colorbox{red!25}{42.0} & 15.6 & \colorbox{blue!15}{54.4} & \colorbox{blue!15}{43.1} \\\midrule
\method{} & \colorbox{red!25}{45.9} & \colorbox{blue!15}{63.3} & \colorbox{red!25}{45.9} & \colorbox{red!25}{43.2} & \colorbox{blue!15}{40.3} & \colorbox{blue!15}{20.2} & \colorbox{red!25}{73.8} & \colorbox{red!25}{47.1} \\
\bottomrule
\end{tabular}
\vspace{-1.5em}
\end{table*}

We evaluate \method{} on nine challenging benchmarks for LLM agents: ALFWorld, WebShop, and seven search-augmented QA tasks. Our experiments address the following questions: 1) How does \method{} compare to state-of-the-art methods? 2) What is the contribution of each component? 3) How does the skill library evolve during training? 4) Does skills accelerate model convergence?

\subsection{Experimental Setup}

\noindent \textbf{Environments.}
ALFWorld \cite{shridharalfworld} is a text-based game aligned with the ALFRED embodied AI benchmark. Agents must complete household tasks by navigating and interacting with objects through text commands. 
WebShop \cite{yao2022webshop} simulates web shopping. Agents navigate a realistic web interface to find and purchase products matching user specifications.  In addition, we also evaluate the performance of \method{} on search-augmented QA tasks, including single-hop QA datasets (NQ~\cite{kwiatkowski2019natural}, TriviaQA~\cite{joshi2017triviaqa}, and PopQA~\cite{mallen2023not}) and multi-hop QA datasets (HotpotQA~\cite{yang2018hotpotqa}, 2Wiki~\cite{ho2020constructing}, MuSiQue~\cite{trivedi2022musique}, and Bamboogle~\cite{press2023measuring}).

\noindent \textbf{Baselines.} We compare \method{} against four categories of competitive methods. First, we include closed-source LLMs, specifically GPT-4o~\cite{openai2024gpt4o} and Gemini-2.5-Pro~\cite{comanici2025gemini}, which represent the state-of-the-art in general-purpose reasoning and instruction following. Second, we evaluate prompt-based agentic or memory-based methods, including ReAct~\cite{yao2022react} and Reflexion~\cite{shinn2023reflexion}, which rely on in-context prompting for multi-step reasoning, as well as Mem0~\cite{chhikara2025mem0}, ExpeL~\cite{zhao2024expel}, and MemP~\cite{fang2025memp}, which utilize external memory or experience pools to guide behavior without parameter updates. Third, we consider RL-based methods, including group-based online RL algorithms such as RLOO~\cite{ahmadian2024back} and GRPO~\cite{shao2024deepseekmath} that optimize policies via advantage estimation over trajectory groups. Finally, we compare against memory-augmented RL-based methods, such as EvolveR~\cite{wu2025evolver}, MemRL~\cite{zhang2026memrl}, and the combination of Mem0+GRPO and SimpleMem~\cite{liu2026simplemem}+GRPO, which integrate persistent memory mechanisms directly into the reinforcement learning optimization process to handle long-term dependencies.
For search-augmented QA, we compare \method{} with R1-Instruct, Search-o1~\cite{li2025search}, Search-R1~\cite{jin2025search}, ZeroSearch~\cite{sun2025zerosearch}, and StepSearch~\cite{zheng2025stepsearch}.

\noindent \textbf{Implementation Details.} We use Qwen2.5-7B-Instruct~\cite{bai2023qwen} as our base model and OpenAI o3~\cite{openai2025o3} as the teacher model for skill distillation and SFT data generation. For RL training, we use GRPO with learning rate $1 \times 10^{-6}$, batch size 16, group size 8, and 4 gradient accumulation steps. We set $K=6$ for task-specific skill retrieval and $\delta=0.4$ for the collection of failed trajectories. For more detailed information on training hyperparameters, please see Appendix~\ref{app:hyp}.

\subsection{Main Results}

\noindent \textbf{Comparison with Baselines.} 
We compare \method{} with baselines across two benchmarks as shown in Table~\ref{tab:performance}. Our method consistently outperforms all baselines, with key observations as follows:

1) \emph{Significant Gains over Prompt-based Methods}. \method{} achieves a 89.9\% success rate on ALFWorld and 72.7\% on WebShop, outperforming the best prompt-based baselines by a large margin. This gap suggests that while in-context learning can leverage past experiences, it often fails to distill actionable knowledge from verbose trajectories or fundamentally adapt the agent's policy.

2) \emph{Superiority over Vanilla RL}. RL training brings substantial gains, yet \method{} consistently surpasses standard RL baselines. Compared to PPO, RLOO, and GRPO, \method{} achieves the best overall performance. Notably, since \method{} utilizes GRPO as its base optimizer, the 12.3\% absolute improvement over GRPO on ALFWorld (from 77.6\% to 89.9\%) is directly attributable to our skill-augmentation mechanism rather than algorithmic variance. In complex subtasks like \textit{Cool} and \textit{Pick2}, \method{} outperforms GRPO by 23.0\% and 22.8\% respectively, proving that structured skill priors effectively accelerate and enhance policy learning in sparse-reward environments.

3) \emph{Advantage over Memory-Augmented RL.} \method{} substantially outperforms existing memory-augmented RL frameworks, which differ in how they manage and update experience. MemRL, which uses RL solely to update its memory bank while keeping the policy frozen, fails to adapt to complex environments, yielding only 21.4\% on ALFWorld. EvolveR, which jointly updates the policy and memory bank, shows improvement (43.8\%) but remains limited by its reliance on rough trajectory storage. To provide a more competitive baseline, we implemented Mem0+GRPO, which combines a state-of-the-art prompt-based memory mechanism with an optimized policy model. While this hybrid approach improves performance to 54.7\% on ALFWorld and 37.5\% on WebShop, it still trails \method{} by a wide margin (about 35.2\% absolute success rate gap). These results validate our core hypothesis: effective experience transfer requires high-level skill abstraction and a co-evolving library rather than simple trajectory compression or prompt-based memory retrieval.

\textbf{Comparison with Closed-Source Models.} Remarkably, \method{} with Qwen2.5-7B-Instruct significantly outperforms much larger closed-source models, as shown in Table~\ref{tab:performance}. On ALFWorld, our method exceeds GPT-4o~\cite{openai2024gpt4o} by 41.9\% and Gemini-2.5-Pro~\cite{comanici2025gemini} by 29.6\%. This demonstrates that effective skill learning can compensate for model scale, enabling smaller open-source models to achieve superior task performance through structured experiential knowledge.

\textbf{Performance on Search-Augmented QA.} As shown in Table~\ref{tab:performance_search}, \method{} achieves a state-of-the-art average score of 47.1\%, significantly outperforming Search-R1 (38.5\%) abd EvolveR (43.1\%). Key observations include: 1) Superior multi-hop Reasoning: \method{} excels in complex tasks like Bamboogle, surpassing EvolveR by 19.4\%. This demonstrates that hierarchical skills effectively guide multi-step information synthesis. 2) Strong generalization: Despite being trained on limited datasets (NQ, HotpotQA), \method{} maintains competitive performance on OOD tasks like TriviaQA and 2Wiki, confirming that distilled search strategies are task-agnostic.

\subsection{Analysis}
In this section, we provide detailed analysis of each module's effectiveness and the skill evolution dynamics. 
\begin{wraptable}{r}{0.3\textwidth}
    \centering
    \vspace{-10pt}
    \caption{Ablation study results. Average success rate (\%) on ALFWorld and WebShop.}
    \label{tab:ablation}
    \resizebox{\linewidth}{!}{ 
        \begin{tabular}{lcc}
            \toprule
            \textbf{Method} & \textbf{ALFWorld} & \textbf{WebShop} \\
            \midrule
            \method{}  & \textbf{89.9} & \textbf{72.7} \\
            \midrule
            \multicolumn{3}{l}{\textit{Skill Library Ablations}} \\
            \quad w/o Hier. & 76.8 & 61.4 \\
            \quad w/o Skill Lib. & 61.7 & 50.2 \\
            \midrule
            \multicolumn{3}{l}{\textit{Training Pipeline Ablations}} \\
            \quad w/o SFT & 65.2 & 46.5 \\
            \quad w/o Dyn. & 84.4 & 70.3 \\
            \bottomrule
        \end{tabular}
    }
    \vspace{-10pt} 
\end{wraptable}

\textbf{Ablation Studies.} We conduct ablation experiments to evaluate each component's contribution, with results in Table~\ref{tab:ablation}. According to the results: (1) Removing hierarchical structure (i.e., task-specific skills only) decreases performance by 13.1\% on ALFWorld and 11.3\% on WebShop, indicating universal strategic principles provide essential foundational guidance. (2) Replacing the skill library with raw trajectories causes the largest degradation (up to 25\%), which directly supports our motivation that abstraction is superior to memorization. Raw experiences introduce significant redundancy and noise that hinder effective knowledge transfer. (3) Cold-start SFT proves critical (20\% drop without it), confirming that the base model requires an initial explicit demonstration phase to learn how to adaptively retrieve and utilize the abstracted skills before entering the RL stage. (4) Dynamic evolution contributes a 5.5\% improvement by ensuring the skill library is a dynamic component rather than a static database. This co-evolution allows the agent to iteratively refine its internal policy by addressing emergent failure modes that were not covered by the initial skill set.

\noindent \textbf{Per-Task Analysis on ALFWorld.} \Cref{tab:performance} breaks down ALFWorld performance by task type. The largest gains are on PickTwo (+23\%), Cool (+22\%) and Heat (+15\%), which are among the most challenging tasks requiring multi-step planning and state tracking. Task-specific skills are particularly valuable here, capturing strategies like ``when picking two objects, verify the first is secured before searching for the second'' that address common failure modes.

\begin{wrapfigure}{r}{0.5\textwidth}
    \centering
    \vspace{-10pt}
    \includegraphics[width=\linewidth]{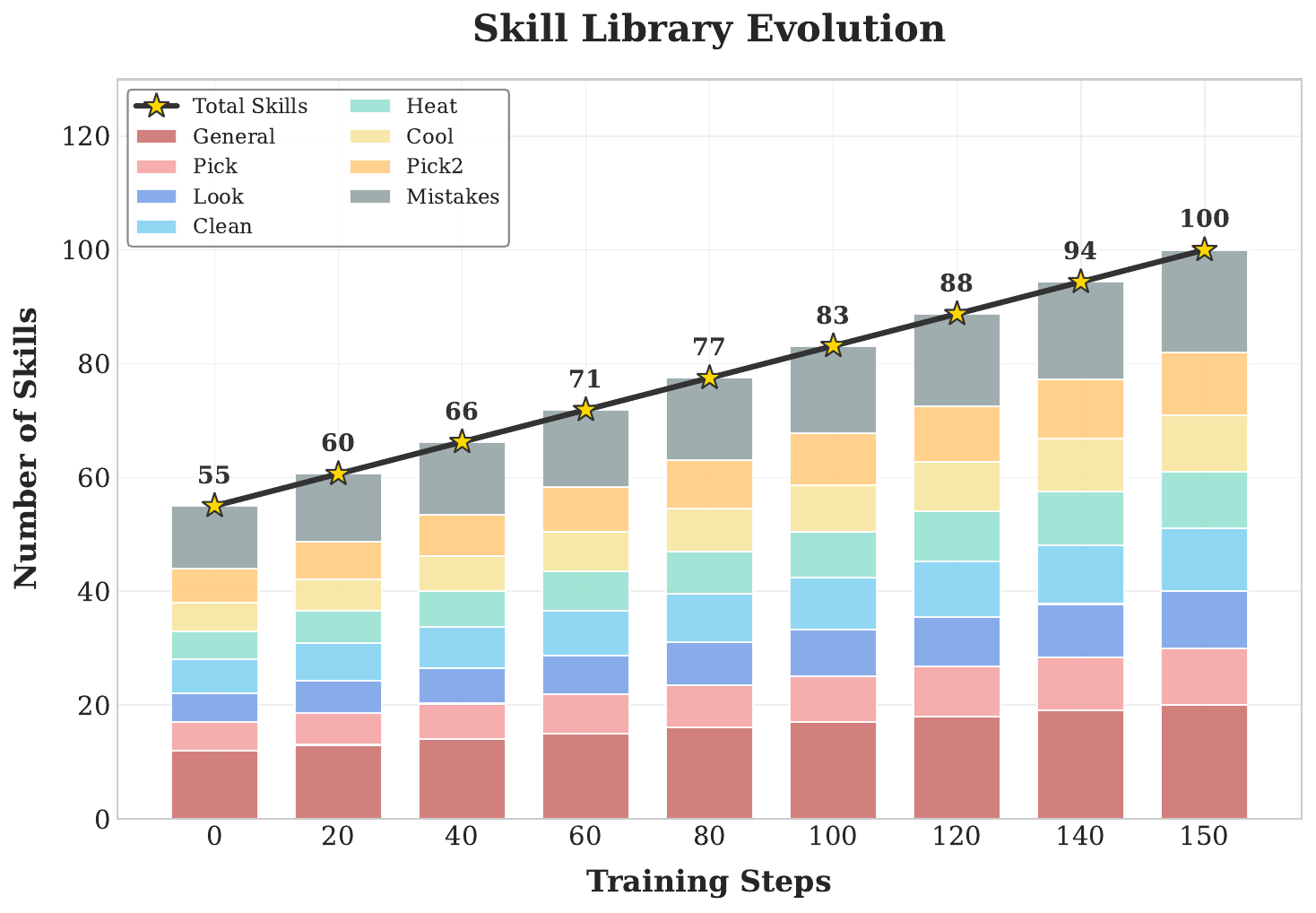}
    \caption{Evolution of skill library size during RL training.}
    \label{fig:skill_evolution}
    \vspace{-10pt}
\end{wrapfigure}

\noindent \textbf{Skill Library Growth.} \Cref{fig:skill_evolution} shows how the skill library evolves during training. The initial skill library contains 55 skills (12 general, 43 task-specific). Through dynamic evolution, this grows to 100 skills by the end of training (Step 150). The growth is predominantly driven by task-specific skills (increasing from 43 to 80), while general skills show a steadier increase (from 12 to 20). Notably, we observe a balanced expansion across various task categories, ensuring the agent develops specialized expertise for each environment rollout. This overall expansion reflects the agent's increasing ability to refine its repertoire and tackle diverse scenarios within specific task types.

\begin{figure*}[t] 
    \centering
    \begin{minipage}{0.49\linewidth}
        \centering
        \includegraphics[width=\textwidth]{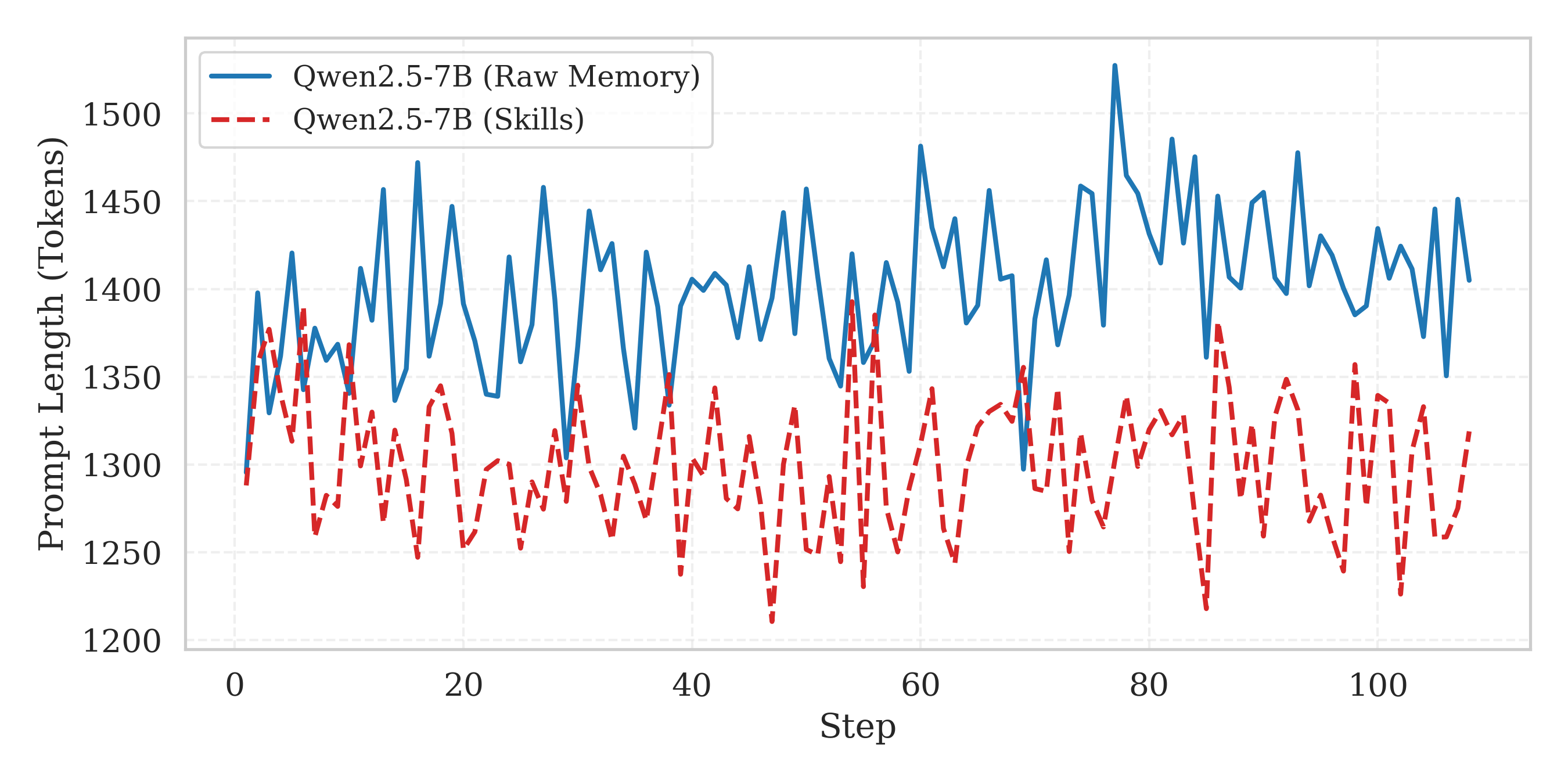}
        \vspace{-1.5em}
        \caption{Comparison of prompt length between raw memory retrieval and our distilled skill abstraction. \method{} consistently reduces context overhead while maintaining reasoning utility.}
        \label{fig:prompt_len}
    \end{minipage}
    \hfill
    \begin{minipage}{0.49\linewidth}
        \centering
        \includegraphics[width=\textwidth]{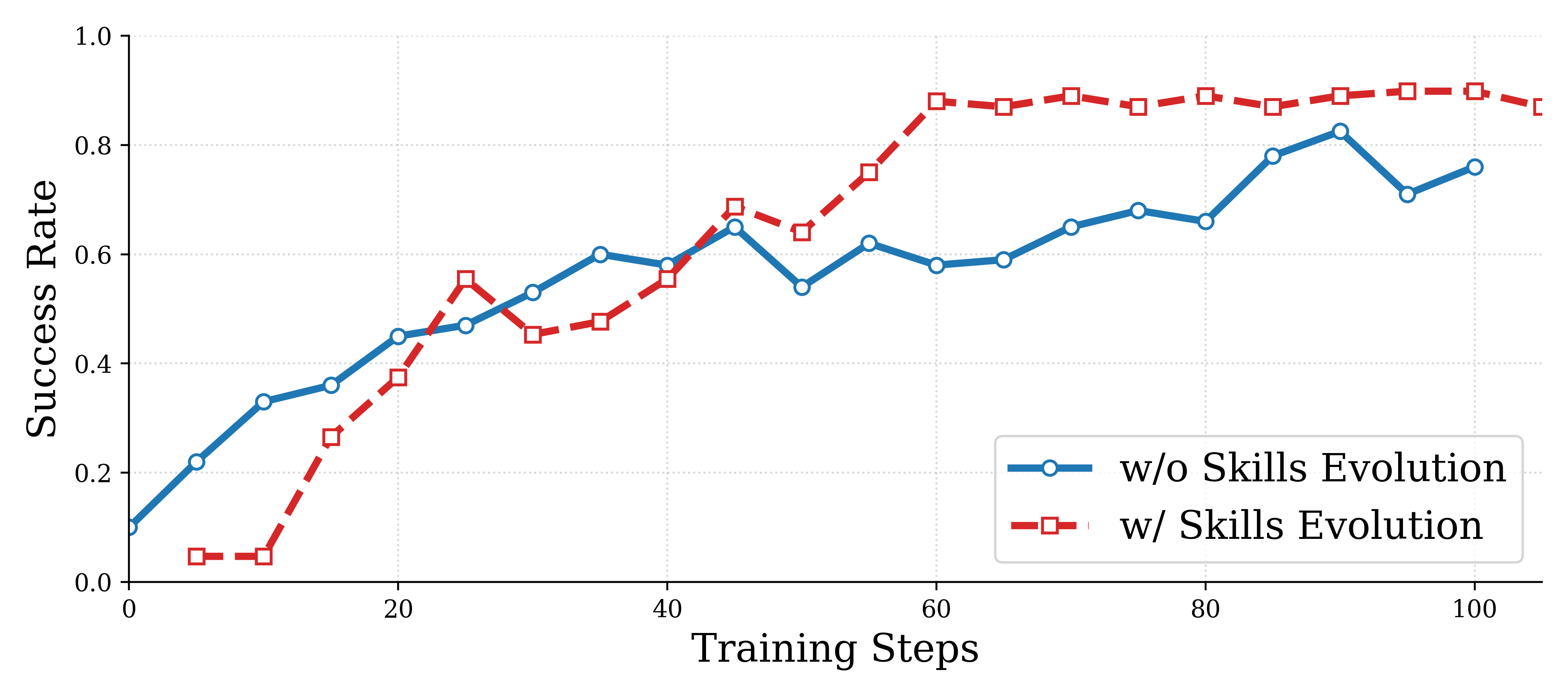}
        \vspace{-1em}
        \caption{Success rate on ALFWorld validation set. The recursive skill evolution significantly accelerates convergence and enhances the overall performance ceiling.}
        \label{fig:sucess_rate}
    \end{minipage}
    \vspace{-2em}
\end{figure*}

\noindent \textbf{Context Efficiency.} 
To evaluate the impact of skill abstraction on inference overhead, we compare the average prompt length of \method{} with a memory-augmented baseline using raw trajectories (Qwen2.5-7B with Raw Memory) in Figure~\ref{fig:prompt_len}. 
The results reveal that while the raw memory approach suffers from a high and fluctuating token footprint (averaging $\sim$1,450 tokens), \method{} maintains a significantly leaner prompt (averaging $<$1,300 tokens), achieving approximately a 10.3\% reduction in context length. 
This efficiency stems from our distillation mechanism, which compresses verbose environment interactions into high-density, actionable skills. 
Notably, \method{} requires less context than the memory-based baseline to achieve superior performance, demonstrating that skill abstraction effectively mitigates the context-bloat problem common in traditional memory-based agents.

\noindent \textbf{Evolution Dynamics.} 
Figure~\ref{fig:sucess_rate} illustrates the reinforcement learning training curves with and without the recursive skill evolution mechanism. 
We observe that while \method{} without evolution shows steady improvement, \method{} with skill evolution exhibits a notably higher learning rate and superior asymptotic performance. 
Specifically, \method{} achieves a success rate of over 80\% within 60 training steps, whereas the baseline requires approximately 90 steps to reach a lower peak. 
This acceleration in convergence suggests that the dynamic introduction of new skills and refinement of existing ones effectively provide the agent with timely strategic guidance to overcome local optima. 
Furthermore, the higher performance ceiling validates that the co-evolution of the skill library and the policy allows the agent to adapt to increasingly complex task scenarios that static memory methods fail to resolve.

\begin{figure*}[t]
    \centering
    \includegraphics[width=0.95\textwidth]{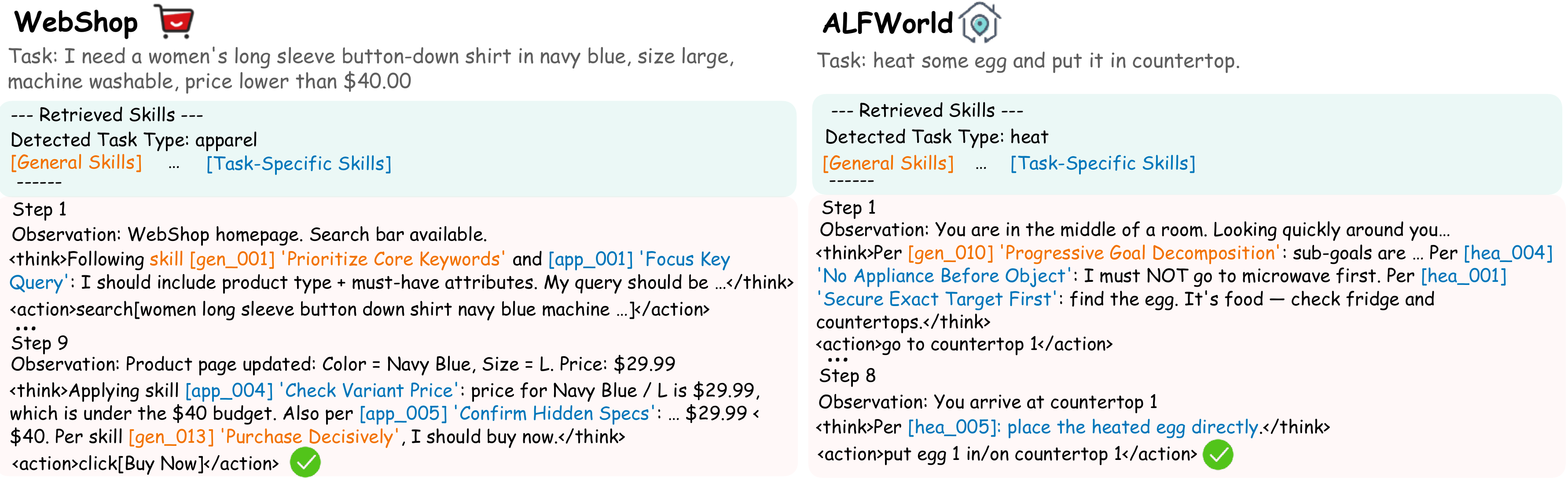}
    \caption{Case studies of \method{} on WebShop and ALFWorld. The examples illustrate how the agent adaptively retrieves and integrates \textcolor[RGB]{255,128,0}{General Skills} and \textcolor[RGB]{0,128,255}{Task-Specific Skills} within its reasoning process to achieve precise and efficient task execution.}
    \label{fig:case}
    \vspace{-0.5em}
\end{figure*}

\noindent \textbf{Qualitative Analysis.}
To further investigate how \method{} utilizes the learned knowledge, we visualize the reasoning process on ALFWorld and WebShop in Figure~\ref{fig:case}. 
The case studies demonstrate that our trained agent can effectively retrieve and execute relevant skills from the \skillbank{} to guide its decision-making. 
For instance, in the WebShop task, the agent invokes general strategies like \textit{``Prioritize Core Keywords''} alongside task-specific heuristics \textit{``Focus Key Query''} to ensure the product meets all constraints within a limited budget. 
Similarly, in ALFWorld, the agent coordinates hierarchical skills, i.e., using \textit{``Progressive Goal Decomposition''} for high-level planning and \textit{``No Appliance Before Object''} to avoid common logical pitfalls. 
This seamless integration of general and specific skills confirms that the agent does not merely memorize trajectories, but rather develops a structured understanding of task logic, allowing for more robust and efficient problem-solving.

\section{Related Work}
\label{sec:related}

\noindent \textbf{LLM Agents.} The emergence of capable LLMs has catalyzed rapid development in autonomous agent systems~\citep{wei2026agentic}. ReAct \cite{yao2022react} interleaves reasoning and acting, enabling chain-of-thought style planning during interaction, while Reflexion \cite{shinn2023reflexion} introduces verbal reinforcement through self-reflection on past failures. Frameworks like AutoGen \cite{wu2024autogen} and CAMEL \cite{li2023camel} demonstrate general-purpose multi-agent capabilities, featuring automated orchestration and diverse tool integration. While initial efforts focused on constrained tasks like coding or basic arithmetic, these approaches primarily rely on in-context learning (ICL)~\cite{dong2024survey}. However, these agents struggle to scale as tasks become more complex, as they treat every interaction as an isolated event and must start each new task from scratch without any prior knowledge.

\noindent \textbf{Memory Mechanisms in Agents.} To overcome the limitations of finite context windows and the inability of agents to learn from experience, external memory architectures have become a cornerstone of agent design~\cite{hu2025memory}. Early systems primarily utilized a static RAG paradigm or stored raw trajectories as few-shot examples~\cite{wangvoyager,chhikara2025mem0,zhang2025g,wang2024agent}. However, raw trajectories are often token-heavy and contain significant redundancy and noise, which can lead to performance degradation. Current research has moved toward self-improving memory, distilling interactions into higher-level insights or procedural tips~\cite{wang2025mirix,tang2025agent,fang2025memp,zhao2024expel,ouyang2025reasoningbank,wei2025evo}. While some recent work explores updating memory banks via online training to improve efficiency~\cite{zhang2025memevolve,zhang2026memrl}, many existing methods still struggle to distinguish high-value experiences from noise or fail to distill core principles that can guide internal decision-making.

\noindent \textbf{Evolution of Agentic Skills and Reinforcement Learning.} The development of agentic skills~\cite{authropic3}, which are compact, reusable strategies that capture the essence of subtasks, is increasingly viewed through the lens of Continual Learning (CL) and RL. Traditional CL~\cite{parisi2019continual} focuses on knowledge preservation in predefined tasks, but self-evolving agents~\cite{gao2025survey,xia2025agent0,liu2025agent0} aim for active skill acquisition in open-ended environments~\cite{fang2025memp,wang2025mem}. While RL is widely used to align LLMs~\cite{schulman2017proximal,ouyang2022training}, or improve reasoning via rule-based verifiers~\cite{shao2024deepseekmath}, applying it to agentic skills remains challenging due to sparse rewards and long horizons. Unlike previous memory-augmented RL which treats memory as a static or auxiliary source, recent trends suggest that the key to efficient experience transfer lies in abstraction~\cite{wu2025evolver}. Our work builds on this by treating the skill library as a dynamic component that co-evolves with the agent's policy, utilizing RL to refine structured skills through recursive failure analysis.
\section{Conclusion}
\label{sec:conclusion}

We introduced \method{}, a framework for skill-augmented reinforcement learning in LLM agents. By distilling raw trajectories into compact, reusable skills and enabling dynamic skill evolution during training, \method{} achieves state-of-the-art performance on ALFWorld and WebShop while using substantially less context than memory-based approaches. Our work demonstrates that the abstraction from experience to skill is a powerful principle for building capable, sample-efficient agents. 

\begin{ack}
This work was partially supported by the Amazon Research Award, the Cisco Faculty Research Award, NEC Laboratories America Research Grant, and Coefficient Giving.

\end{ack}

\bibliographystyle{plainnat}
\bibliography{main}

@inproceedings{yao2022react,
  title={React: Synergizing reasoning and acting in language models},
  author={Yao, Shunyu and Zhao, Jeffrey and Yu, Dian and Du, Nan and Shafran, Izhak and Narasimhan, Karthik R and Cao, Yuan},
  booktitle={The eleventh international conference on learning representations},
  year={2022}
}

@article{xia2026rrsi,
  title={RRSI: Regularized Recursive Self-Improvement of Agent Harnesses},
  author={Xia, Peng and Han, Rujun and Wang, Zifeng and Chen, Yanfei and Zhang, Yufan and Lee, Yoonho and Huang, Chengsong and Yu, Han and CuiZhu, Zhongying and Ming, Yifei and others},
  journal={arXiv preprint arXiv:2609.24972},
  year={2026}
}

@article{huang2026envharness,
  title={EnvHarness: Awakening static worlds for agent learning},
  author={Huang, Chengsong and Wang, Zifeng and Han, Rujun and Yan, Jun and Chen, Yanfei and CuiZhu, Zoey and Jiang, Ke and Xia, Peng and Yu, Han and Zhuang, Yufan and others},
  journal={arXiv preprint arXiv:2608.19880},
  year={2026}
}

@misc{tinker,
  author={Thinking Machines Lab TML},
  title={Tinker is a training API for researchers},
  note={\url{https://thinkingmachines.ai/tinker/}},
  year={2026}
}

@article{team2026kimi,
  title={Kimi K2. 5: Visual Agentic Intelligence},
  author={Team, Kimi and Bai, Tongtong and Bai, Yifan and Bao, Yiping and Cai, SH and Cao, Yuan and Charles, Y and Che, HS and Chen, Cheng and Chen, Guanduo and others},
  journal={arXiv preprint arXiv:2602.02276},
  year={2026}
}

@article{li2025search,
  title={Search-o1: Agentic search-enhanced large reasoning models},
  author={Li, Xiaoxi and Dong, Guanting and Jin, Jiajie and Zhang, Yuyao and Zhou, Yujia and Zhu, Yutao and Zhang, Peitian and Dou, Zhicheng},
  journal={arXiv preprint arXiv:2501.05366},
  year={2025}
}

@inproceedings{zheng2025stepsearch,
  title={StepSearch: Igniting LLMs search ability via step-wise proximal policy optimization},
  author={Zheng, Xuhui and An, Kang and Wang, Ziliang and Wang, Yuhang and Wu, Yichao},
  booktitle={Proceedings of the 2025 Conference on Empirical Methods in Natural Language Processing},
  pages={21816--21841},
  year={2025}
}

@article{jin2025search,
  title={Search-r1: Training llms to reason and leverage search engines with reinforcement learning},
  author={Jin, Bowen and Zeng, Hansi and Yue, Zhenrui and Yoon, Jinsung and Arik, Sercan and Wang, Dong and Zamani, Hamed and Han, Jiawei},
  journal={arXiv preprint arXiv:2503.09516},
  year={2025}
}

@inproceedings{press2023measuring,
  title={Measuring and narrowing the compositionality gap in language models},
  author={Press, Ofir and Zhang, Muru and Min, Sewon and Schmidt, Ludwig and Smith, Noah A and Lewis, Mike},
  booktitle={Findings of the Association for Computational Linguistics: EMNLP 2023},
  pages={5687--5711},
  year={2023}
}

@article{trivedi2022musique,
  title={MuSiQue: Multihop Questions via Single-hop Question Composition},
  author={Trivedi, Harsh and Balasubramanian, Niranjan and Khot, Tushar and Sabharwal, Ashish},
  journal={Transactions of the Association for Computational Linguistics},
  volume={10},
  pages={539--554},
  year={2022},
  publisher={MIT Press One Broadway, 12th Floor, Cambridge, Massachusetts 02142, USA~…}
}

@inproceedings{ho2020constructing,
  title={Constructing A Multi-hop QA Dataset for Comprehensive Evaluation of Reasoning Steps},
  author={Ho, Xanh and Nguyen, Anh-Khoa Duong and Sugawara, Saku and Aizawa, Akiko},
  booktitle={Proceedings of the 28th International Conference on Computational Linguistics},
  pages={6609--6625},
  year={2020}
}

@inproceedings{yang2018hotpotqa,
  title={HotpotQA: A dataset for diverse, explainable multi-hop question answering},
  author={Yang, Zhilin and Qi, Peng and Zhang, Saizheng and Bengio, Yoshua and Cohen, William and Salakhutdinov, Ruslan and Manning, Christopher D},
  booktitle={Proceedings of the 2018 conference on empirical methods in natural language processing},
  pages={2369--2380},
  year={2018}
}

@inproceedings{mallen2023not,
  title={When not to trust language models: Investigating effectiveness of parametric and non-parametric memories},
  author={Mallen, Alex and Asai, Akari and Zhong, Victor and Das, Rajarshi and Khashabi, Daniel and Hajishirzi, Hannaneh},
  booktitle={Proceedings of the 61st Annual Meeting of the Association for Computational Linguistics (Volume 1: Long Papers)},
  pages={9802--9822},
  year={2023}
}

@inproceedings{joshi2017triviaqa,
  title={TriviaQA: A Large Scale Distantly Supervised Challenge Dataset for Reading Comprehension},
  author={Joshi, Mandar and Choi, Eunsol and Weld, Daniel S and Zettlemoyer, Luke},
  booktitle={Proceedings of the 55th Annual Meeting of the Association for Computational Linguistics (Volume 1: Long Papers)},
  pages={1601--1611},
  year={2017}
}

@article{zhang2025memevolve,
  title={MemEvolve: Meta-Evolution of Agent Memory Systems},
  author={Zhang, Guibin and Ren, Haotian and Zhan, Chong and Zhou, Zhenhong and Wang, Junhao and Zhu, He and Zhou, Wangchunshu and Yan, Shuicheng},
  journal={arXiv preprint arXiv:2512.18746},
  year={2025}
}

@article{kwiatkowski2019natural,
  title={Natural questions: a benchmark for question answering research},
  author={Kwiatkowski, Tom and Palomaki, Jennimaria and Redfield, Olivia and Collins, Michael and Parikh, Ankur and Alberti, Chris and Epstein, Danielle and Polosukhin, Illia and Devlin, Jacob and Lee, Kenton and others},
  journal={Transactions of the Association for Computational Linguistics},
  volume={7},
  pages={453--466},
  year={2019},
  publisher={MIT Press One Rogers Street, Cambridge, MA 02142-1209, USA journals-info~…}
}

@article{sun2025zerosearch,
  title={Zerosearch: Incentivize the search capability of llms without searching},
  author={Sun, Hao and Qiao, Zile and Guo, Jiayan and Fan, Xuanbo and Hou, Yingyan and Jiang, Yong and Xie, Pengjun and Zhang, Yan and Huang, Fei and Zhou, Jingren},
  journal={arXiv preprint arXiv:2505.04588},
  year={2025}
}

@article{wang2025mirix,
  title={Mirix: Multi-agent memory system for llm-based agents},
  author={Wang, Yu and Chen, Xi},
  journal={arXiv preprint arXiv:2507.07957},
  year={2025}
}

@article{wei2026agentic,
  title={Agentic Reasoning for Large Language Models},
  author={Wei, Tianxin and Li, Ting-Wei and Liu, Zhining and Ning, Xuying and Yang, Ze and Zou, Jiaru and Zeng, Zhichen and Qiu, Ruizhong and Lin, Xiao and Fu, Dongqi and others},
  journal={arXiv preprint arXiv:2601.12538},
  year={2026}
}

@article{wei2025evo,
  title={Evo-Memory: Benchmarking LLM Agent Test-time Learning with Self-Evolving Memory},
  author={Wei, Tianxin and Sachdeva, Noveen and Coleman, Benjamin and He, Zhankui and Bei, Yuanchen and Ning, Xuying and Ai, Mengting and Li, Yunzhe and He, Jingrui and Chi, Ed H and others},
  journal={arXiv preprint arXiv:2511.20857},
  year={2025}
}

@article{liu2026simplemem,
  title={SimpleMem: Efficient Lifelong Memory for LLM Agents},
  author={Liu, Jiaqi and Su, Yaofeng and Xia, Peng and Han, Siwei and Zheng, Zeyu and Xie, Cihang and Ding, Mingyu and Yao, Huaxiu},
  journal={arXiv preprint arXiv:2601.02553},
  year={2026}
}

@article{zhang2026memrl,
  title={MemRL: Self-Evolving Agents via Runtime Reinforcement Learning on Episodic Memory},
  author={Zhang, Shengtao and Wang, Jiaqian and Zhou, Ruiwen and Liao, Junwei and Feng, Yuchen and Zhang, Weinan and Wen, Ying and Li, Zhiyu and Xiong, Feiyu and Qi, Yutao and others},
  journal={arXiv preprint arXiv:2601.03192},
  year={2026}
}

@article{wang2025mem,
  title={Mem-$\{$$\backslash$alpha$\}$: Learning memory construction via reinforcement learning},
  author={Wang, Yu and Takanobu, Ryuichi and Liang, Zhiqi and Mao, Yuzhen and Hu, Yuanzhe and McAuley, Julian and Wu, Xiaojian},
  journal={arXiv preprint arXiv:2509.25911},
  year={2025}
}

@inproceedings{dong2024survey,
  title={A survey on in-context learning},
  author={Dong, Qingxiu and Li, Lei and Dai, Damai and Zheng, Ce and Ma, Jingyuan and Li, Rui and Xia, Heming and Xu, Jingjing and Wu, Zhiyong and Chang, Baobao and others},
  booktitle={Proceedings of the 2024 conference on empirical methods in natural language processing},
  pages={1107--1128},
  year={2024}
}

@article{wang2024agent,
  title={Agent workflow memory},
  author={Wang, Zora Zhiruo and Mao, Jiayuan and Fried, Daniel and Neubig, Graham},
  journal={arXiv preprint arXiv:2409.07429},
  year={2024}
}

@article{zhang2025g,
  title={G-Memory: Tracing Hierarchical Memory for Multi-Agent Systems},
  author={Zhang, Guibin and Fu, Muxin and Wan, Guancheng and Yu, Miao and Wang, Kun and Yan, Shuicheng},
  journal={arXiv preprint arXiv:2506.07398},
  year={2025}
}

@article{ouyang2022training,
  title={Training language models to follow instructions with human feedback},
  author={Ouyang, Long and Wu, Jeffrey and Jiang, Xu and Almeida, Diogo and Wainwright, Carroll and Mishkin, Pamela and Zhang, Chong and Agarwal, Sandhini and Slama, Katarina and Ray, Alex and others},
  journal={Advances in neural information processing systems},
  volume={35},
  pages={27730--27744},
  year={2022}
}

@article{gao2025survey,
  title={A survey of self-evolving agents: On path to artificial super intelligence},
  author={Gao, Huan-ang and Geng, Jiayi and Hua, Wenyue and Hu, Mengkang and Juan, Xinzhe and Liu, Hongzhang and Liu, Shilong and Qiu, Jiahao and Qi, Xuan and Wu, Yiran and others},
  journal={arXiv preprint arXiv:2507.21046},
  year={2025}
}

@article{xia2025agent0,
  title={Agent0: Unleashing self-evolving agents from zero data via tool-integrated reasoning},
  author={Xia, Peng and Zeng, Kaide and Liu, Jiaqi and Qin, Can and Wu, Fang and Zhou, Yiyang and Xiong, Caiming and Yao, Huaxiu},
  journal={arXiv preprint arXiv:2511.16043},
  year={2025}
}

@article{liu2025agent0,
  title={Agent0-VL: Exploring Self-Evolving Agent for Tool-Integrated Vision-Language Reasoning},
  author={Liu, Jiaqi and Xiong, Kaiwen and Xia, Peng and Zhou, Yiyang and Ji, Haonian and Feng, Lu and Han, Siwei and Ding, Mingyu and Yao, Huaxiu},
  journal={arXiv preprint arXiv:2511.19900},
  year={2025}
}

@article{parisi2019continual,
  title={Continual lifelong learning with neural networks: A review},
  author={Parisi, German I and Kemker, Ronald and Part, Jose L and Kanan, Christopher and Wermter, Stefan},
  journal={Neural networks},
  volume={113},
  pages={54--71},
  year={2019},
  publisher={Elsevier}
}

@article{hu2025memory,
  title={Memory in the Age of AI Agents},
  author={Hu, Yuyang and Liu, Shichun and Yue, Yanwei and Zhang, Guibin and Liu, Boyang and Zhu, Fangyi and Lin, Jiahang and Guo, Honglin and Dou, Shihan and Xi, Zhiheng and others},
  journal={arXiv preprint arXiv:2512.13564},
  year={2025}
}

@article{ouyang2025reasoningbank,
  title={Reasoningbank: Scaling agent self-evolving with reasoning memory},
  author={Ouyang, Siru and Yan, Jun and Hsu, I and Chen, Yanfei and Jiang, Ke and Wang, Zifeng and Han, Rujun and Le, Long T and Daruki, Samira and Tang, Xiangru and others},
  journal={arXiv preprint arXiv:2509.25140},
  year={2025}
}

@article{tang2025agent,
  title={Agent kb: Leveraging cross-domain experience for agentic problem solving},
  author={Tang, Xiangru and Qin, Tianrui and Peng, Tianhao and Zhou, Ziyang and Shao, Daniel and Du, Tingting and Wei, Xinming and Xia, Peng and Wu, Fang and Zhu, He and others},
  journal={arXiv preprint arXiv:2507.06229},
  year={2025}
}

@article{li2023camel,
  title={Camel: Communicative agents for" mind" exploration of large language model society},
  author={Li, Guohao and Hammoud, Hasan and Itani, Hani and Khizbullin, Dmitrii and Ghanem, Bernard},
  journal={Advances in Neural Information Processing Systems},
  volume={36},
  pages={51991--52008},
  year={2023}
}

@inproceedings{ahmadian2024back,
  title={Back to Basics: Revisiting REINFORCE-Style Optimization for Learning from Human Feedback in LLMs},
  author={Ahmadian, Arash and Cremer, Chris and Gall{\'e}, Matthias and Fadaee, Marzieh and Kreutzer, Julia and Pietquin, Olivier and {\"U}st{\"u}n, Ahmet and Hooker, Sara},
  booktitle={Proceedings of the 62nd Annual Meeting of the Association for Computational Linguistics (Volume 1: Long Papers)},
  pages={12248--12267},
  year={2024}
}

@article{comanici2025gemini,
  title={Gemini 2.5: Pushing the frontier with advanced reasoning, multimodality, long context, and next generation agentic capabilities},
  author={Comanici, Gheorghe and Bieber, Eric and Schaekermann, Mike and Pasupat, Ice and Sachdeva, Noveen and Dhillon, Inderjit and Blistein, Marcel and Ram, Ori and Zhang, Dan and Rosen, Evan and others},
  journal={arXiv preprint arXiv:2507.06261},
  year={2025}
}

@article{wu2025evolver,
  title={Evolver: Self-evolving llm agents through an experience-driven lifecycle},
  author={Wu, Rong and Wang, Xiaoman and Mei, Jianbiao and Cai, Pinlong and Fu, Daocheng and Yang, Cheng and Wen, Licheng and Yang, Xuemeng and Shen, Yufan and Wang, Yuxin and others},
  journal={arXiv preprint arXiv:2510.16079},
  year={2025}
}

@misc{openai2024gpt4o,
  author={OpenAI},
  title={GPT-4o System Card},
  note={\url{https://openai.com/index/gpt-4o-system-card/}},
  year={2024}
}

@article{fang2025memp,
  title={Memp: Exploring agent procedural memory},
  author={Fang, Runnan and Liang, Yuan and Wang, Xiaobin and Wu, Jialong and Qiao, Shuofei and Xie, Pengjun and Huang, Fei and Chen, Huajun and Zhang, Ningyu},
  journal={arXiv preprint arXiv:2508.06433},
  year={2025}
}

@article{guo2025deepseek,
  title={Deepseek-r1: Incentivizing reasoning capability in llms via reinforcement learning},
  author={Guo, Daya and Yang, Dejian and Zhang, Haowei and Song, Junxiao and Zhang, Ruoyu and Xu, Runxin and Zhu, Qihao and Ma, Shirong and Wang, Peiyi and Bi, Xiao and others},
  journal={arXiv preprint arXiv:2501.12948},
  year={2025}
}

@article{feng2025group,
  title={Group-in-group policy optimization for llm agent training},
  author={Feng, Lang and Xue, Zhenghai and Liu, Tingcong and An, Bo},
  journal={arXiv preprint arXiv:2505.10978},
  year={2025}
}

@misc{openaigui,
title={OpenAI Computer-Using Agent},
url={https://openai.com/index/computer-using-agent/},
year={2025},
author={OpenAI}
}

@article{shao2024deepseekmath,
  title={Deepseekmath: Pushing the limits of mathematical reasoning in open language models},
  author={Shao, Zhihong and Wang, Peiyi and Zhu, Qihao and Xu, Runxin and Song, Junxiao and Bi, Xiao and Zhang, Haowei and Zhang, Mingchuan and Li, YK and Wu, Yang and others},
  journal={arXiv preprint arXiv:2402.03300},
  year={2024}
}

@misc{authropic3,
title={The claude 3 model family: Opus, sonnet, haiku},
url={https://www.anthropic.com/news/claude-3-family},
year={2024},
author={Anthropic}
}

@misc{geminigui,
title={Introducing the Gemini 2.5 Computer Use model},
url={https://blog.google/technology/google-deepmind/gemini-computer-use-model/},
year={2025},
author={Google}
}

@misc{openaideepresearch,
title={OpenAI Deep Research System Card},
url={https://openai.com/index/introducing-deep-research/},
year={2025},
author={OpenAI}
}

@misc{geminideepresearch,
title={Try Deep Research and our new experimental model in Gemini, your AI assistant},
url={https://blog.google/products/gemini/google-gemini-deep-research/},
year={2024},
author={Google}
}

@article{team2025tongyi,
  title={Tongyi DeepResearch Technical Report},
  author={Team, Tongyi DeepResearch and Li, Baixuan and Zhang, Bo and Zhang, Dingchu and Huang, Fei and Li, Guangyu and Chen, Guoxin and Yin, Huifeng and Wu, Jialong and Zhou, Jingren and others},
  journal={arXiv preprint arXiv:2510.24701},
  year={2025}
}

@article{chhikara2025mem0,
  title={Mem0: Building production-ready ai agents with scalable long-term memory},
  author={Chhikara, Prateek and Khant, Dev and Aryan, Saket and Singh, Taranjeet and Yadav, Deshraj},
  journal={arXiv preprint arXiv:2504.19413},
  year={2025}
}

@article{schulman2017proximal,
  title={Proximal policy optimization algorithms},
  author={Schulman, John and Wolski, Filip and Dhariwal, Prafulla and Radford, Alec and Klimov, Oleg},
  journal={arXiv preprint arXiv:1707.06347},
  year={2017}
}

@misc{openai2025o3,
  author={OpenAI},
  title={Introducing o3 and o4-mini},
  note={\url{https://openai.com/index/introducing-o3-and-o4-mini/}},
  year={2025}
}

@article{bai2023qwen,
  title={Qwen technical report},
  author={Bai, Jinze and Bai, Shuai and Chu, Yunfei and Cui, Zeyu and Dang, Kai and Deng, Xiaodong and Fan, Yang and Ge, Wenbin and Han, Yu and Huang, Fei and others},
  journal={arXiv preprint arXiv:2309.16609},
  year={2023}
}

@article{yao2022webshop,
  title={Webshop: Towards scalable real-world web interaction with grounded language agents},
  author={Yao, Shunyu and Chen, Howard and Yang, John and Narasimhan, Karthik},
  journal={Advances in Neural Information Processing Systems},
  volume={35},
  pages={20744--20757},
  year={2022}
}

@inproceedings{shridharalfworld,
  title={ALFWorld: Aligning Text and Embodied Environments for Interactive Learning},
  author={Shridhar, Mohit and Yuan, Xingdi and Cote, Marc-Alexandre and Bisk, Yonatan and Trischler, Adam and Hausknecht, Matthew},
  booktitle={International Conference on Learning Representations}
}

@inproceedings{zhao2024expel,
  title={Expel: Llm agents are experiential learners},
  author={Zhao, Andrew and Huang, Daniel and Xu, Quentin and Lin, Matthieu and Liu, Yong-Jin and Huang, Gao},
  booktitle={Proceedings of the AAAI Conference on Artificial Intelligence},
  volume={38},
  number={17},
  pages={19632--19642},
  year={2024}
}

@article{wangvoyager,
  title={Voyager: An Open-Ended Embodied Agent with Large Language Models},
  author={Wang, Guanzhi and Xie, Yuqi and Jiang, Yunfan and Mandlekar, Ajay and Xiao, Chaowei and Zhu, Yuke and Fan, Linxi and Anandkumar, Anima},
  journal={Transactions on Machine Learning Research}
}

@inproceedings{wu2024autogen,
  title={Autogen: Enabling next-gen LLM applications via multi-agent conversations},
  author={Wu, Qingyun and Bansal, Gagan and Zhang, Jieyu and Wu, Yiran and Li, Beibin and Zhu, Erkang and Jiang, Li and Zhang, Xiaoyun and Zhang, Shaokun and Liu, Jiale and others},
  booktitle={First Conference on Language Modeling},
  year={2024}
}

@article{shinn2023reflexion,
  title={Reflexion: Language agents with verbal reinforcement learning},
  author={Shinn, Noah and Cassano, Federico and Gopinath, Ashwin and Narasimhan, Karthik and Yao, Shunyu},
  journal={Advances in Neural Information Processing Systems},
  volume={36},
  pages={8634--8652},
  year={2023}
}


\newpage
\appendix
\onecolumn
\startcontents[appendix]
\printcontents[appendix]{ }{0}{\section*{Appendix}}

\newpage

\section{Limitations}
\method{} currently relies on a high-capability teacher model for the critical skill distillation and synthetic trajectory generation phases. Additionally, the reinforcement learning pipeline involves substantial computational resources. As the agent operates in increasingly open-ended environments, managing the scale of the growing SkillBank without introducing retrieval noise remains an insightful challenge. Future research could explore transitioning from periodic validation-triggered updates to a more fluid, real-time evolution of skills. Ultimately, expanding these principles to even broader domains like professional software engineering will further test and refine the framework's generalizability.  

\section{Algorithm}
The algorithm is shown in Algorithm~\ref{alg:skillrl}.
\begin{algorithm}[t]
\small
\caption{\method{}: Recursive Skill-Augmented RL}
\label{alg:skillrl}
\begin{algorithmic}[1]
    \REQUIRE Base model $\pi_{\text{base}}$, teacher $\mathcal{M}_T$, environment $\mathcal{E}$
    \ENSURE Trained policy $\pi_{\theta^*}$, evolved skill library $\skillbank{}^*$
    
    \STATE \textcolor{blue}{$\triangleright$ Experience-based Skill Distillation}
    \STATE $\mathcal{T}^+, \mathcal{T}^- \leftarrow \text{Rollout}(\pi_{\text{base}}, \mathcal{E})$
    \FORALL{$\tau^+ \in \mathcal{T}^+$}
        \STATE $s^+ \leftarrow \mathcal{M}_T(\tau^+)$
    \ENDFOR
    \FORALL{$\tau^- \in \mathcal{T}^-$}
        \STATE $s^- \leftarrow \mathcal{M}_T(\tau^-)$
    \ENDFOR
    
    \STATE \textcolor{blue}{$\triangleright$ Hierarchical Skill Library Construction}
    \STATE $\mathcal{S}_g \leftarrow$ general skills from distilled experiences
    \FORALL{task type $k$}
        \STATE $\mathcal{S}_k \leftarrow$ task-specific skills for category $k$
    \ENDFOR
    \STATE $\skillbank{} \leftarrow \mathcal{S}_g \cup \bigcup_k \mathcal{S}_k$
    
    \STATE \textcolor{blue}{$\triangleright$ Recursive Skill Evolution via RL}
    \STATE \textcolor{gray}{\textit{// Cold-start initialization}}
    \STATE $\mathcal{D}_{\text{SFT}} \leftarrow \mathcal{M}_T(\mathcal{E}, \skillbank{})$
    \STATE $\theta \leftarrow \text{SFT}(\pi_{\text{base}}, \mathcal{D}_{\text{SFT}})$; \quad $\pi_{\text{ref}} \leftarrow \pi_\theta$
    \STATE \textcolor{gray}{\textit{// RL with recursive evolution}}
    \FOR{epoch $= 1$ to $N$}
        \FORALL{task $d$}
            \STATE $\mathcal{S}_{\text{ret}} \leftarrow \text{Retrieve}(d, \skillbank{})$
            \STATE Sample $\{\tau^{(i)}\}_{i=1}^G \sim \pi_\theta(\cdot | d, \mathcal{S}_g, \mathcal{S}_{\text{ret}})$
            \STATE Compute $\{R_i\}_{i=1}^G$ and update $\theta$ via GRPO
        \ENDFOR
        \IF{validation epoch}
            \STATE $\mathcal{T}_{\text{val}}^- \leftarrow$ failed validation trajectories
            \STATE $\mathcal{S}_{\text{new}} \leftarrow \mathcal{M}_T(\mathcal{T}_{\text{val}}^-, \skillbank{})$
            \STATE $\skillbank{} \leftarrow \skillbank{} \cup \mathcal{S}_{\text{new}}$
        \ENDIF
    \ENDFOR
    \STATE \textbf{return} $\pi_\theta$, $\skillbank{}$
\end{algorithmic}
\end{algorithm}

\section{Prompts}
\label{app:prompts}
In this section, we provide the full prompt templates used throughout the different phases of our framework. These templates are designed to ensure consistent agent behavior and structured data generation across various environments.

\subsection{Agent Execution Prompts}
The following prompts are used during the online inference phase. These templates provide the agent with the current task description, a history of previous interactions, and a set of retrieved skills (experiences) to guide its decision-making process. The prompts explicitly enforce a Chain-of-Thought (CoT) reasoning step before action selection.
\begin{table}[H]
\centering
\small
\begin{tabularx}{\textwidth}{|X|}
\hline
\rowcolor[gray]{0.9} \textbf{Prompt A.1: ALFWorld Agent Execution with Skills} \\ \hline
\textbf{System Prompt:} \\
You are an expert agent operating in the ALFRED Embodied Environment. Your task is to: \texttt{\{task\_description\}} \\
\\
\textbf{\#\# Retrieved Relevant Experience} \\
\texttt{\{retrieved\_memories\}} \\
\\
\textbf{\#\# Current Progress} \\
Prior to this step, you have already taken \texttt{\{step\_count\}} step(s). Below are the most recent \texttt{\{history\_length\}} observations and the corresponding actions you took: \texttt{\{action\_history\}} \\
You are now at step \texttt{\{current\_step\}} and your current observation is: \texttt{\{current\_observation\}} \\
Your admissible actions of the current situation are: [\texttt{\{admissible\_actions\}}]. \\
\\
Now it's your turn to take an action. You should first reason step-by-step about the current situation. This reasoning process \textbf{MUST} be enclosed within \texttt{<think> </think>} tags. Once you've finished your reasoning, you should choose an admissible action for current step and present it within \texttt{<action> </action>} tags. \\ \hline
\end{tabularx}
\end{table}

\begin{table}[H]
\centering
\small
\begin{tabularx}{\textwidth}{|X|}
\hline
\rowcolor[gray]{0.9} \textbf{Prompt A.2: WebShop Agent Execution with Skills} \\ \hline
\textbf{System Prompt:} \\
You are an expert autonomous agent operating in the WebShop e-commerce environment. Your task is to: \texttt{\{task\_description\}}. \\
\\
\textbf{\#\# Retrieved Relevant Experience} \\
\texttt{\{retrieved\_memories\}} \\
\\
\textbf{\#\# Current Progress} \\
Prior to this step, you have already taken \texttt{\{step\_count\}} step(s). Below are the most recent \texttt{\{history\_length\}} observations and the corresponding actions you took: \texttt{\{action\_history\}} \\
You are now at step \texttt{\{current\_step\}} and your current observation is: \texttt{\{current\_observation\}} \\
Your admissible actions of the current situation are: [ \texttt{\{available\_actions\}} ]. \\
\\
Now it's your turn to take one action for the current step. You should first reason step-by-step about the current situation, then think carefully which admissible action best advances the shopping goal. This reasoning process \textbf{MUST} be enclosed within \texttt{<think> </think>} tags. Once you've finished your reasoning, you should choose an admissible action for current step and present it within \texttt{<action> </action>} tags. \\ \hline
\end{tabularx}
\end{table}

\subsection{Skill Generation and Distillation Prompts}
These prompts are utilized during the skill discovery and library initialization phases. They guide a high-capability teacher model to analyze interaction trajectories, identify failure modes, and distill reusable, actionable skills into a structured JSON format.
\begin{table}[H]
\centering
\small
\begin{tabularx}{\textwidth}{|X|}
\hline
\rowcolor[gray]{0.9} \textbf{Prompt B.1: Dynamic Skill Discovery from Failures} \\ \hline
Analyze these failed \texttt{\{env\_description\}} agent trajectories and suggest NEW skills to add. \\
\\
\textbf{FAILED TRAJECTORIES:} \texttt{\{failure\_examples\}} \\
\textbf{EXISTING SKILL TITLES:} \texttt{\{existing\_titles\}} \\
\\
Generate 1-3 NEW actionable skills that would help avoid these failures. Each skill must have: \texttt{skill\_id}, \texttt{title} (3-5 words), \texttt{principle} (1-2 sentences), \texttt{when\_to\_apply}. The \texttt{skill\_id} should be unique and follow the pattern: "dyn\_001", "dyn\_002", etc. \\
\\
Return ONLY a JSON array of skills, no other text. \\ \hline
\end{tabularx}
\end{table}

\begin{table}[H]
\centering
\small
\begin{tabularx}{\textwidth}{|X|}
\hline
\rowcolor[gray]{0.9} \textbf{Prompt B.2: Initial Skill Distillation (ALFWorld)} \\ \hline
You are an expert at distilling agent behavior patterns into concise, actionable skills. Analyze these successful and failed trajectories from an embodied AI agent operating in household environments (ALFWorld). \\
\\
\textbf{SUCCESSFUL TRAJECTORIES:} \texttt{\{success\_patterns\}} \\
\textbf{FAILED TRAJECTORIES:} \texttt{\{failure\_patterns\}} \\
\\
Generate 8-12 GENERAL SKILLS that apply across ALL task types. These should be: 1. \textbf{Concise}; 2. \textbf{Actionable}; 3. \textbf{Transferable}; 4. \textbf{Failure-aware}. Focus on: Navigation, object manipulation, state tracking, error recovery, and container interaction rules. \\
\\
Return ONLY the JSON array, no other text. \\ \hline
\end{tabularx}
\end{table}

\begin{table}[H]
\centering
\small
\begin{tabularx}{\textwidth}{|X|}
\hline
\rowcolor[gray]{0.9} \textbf{Prompt B.3: Initial Skill Distillation (WebShop)} \\ \hline
You are an expert at distilling agent behavior patterns into concise, actionable skills. Analyze these successful and failed trajectories from an AI agent operating in an online shopping environment (WebShop). \\
\\
\textbf{SUCCESSFUL TRAJECTORIES:} \texttt{\{success\_patterns\}} \\
\textbf{FAILED TRAJECTORIES:} \texttt{\{failure\_patterns\}} \\
\\
Generate 10-15 GENERAL SKILLS. Focus on: Search query formulation, product selection heuristics, option configuration (size, color, etc.), constraint verification, navigation patterns, and price handling. \\
\\
Return ONLY the JSON array, no other text. \\ \hline
\end{tabularx}
\end{table}

\subsection{Cold-start Trajectory Generation Prompts}
To bridge the gap between a base model and the target performance, we use the following prompts to generate high-quality synthetic trajectories for Supervised Fine-Tuning (SFT). These prompts instruct the teacher model to solve tasks while explicitly demonstrating the application of specific skills, thereby providing a clear learning signal for the student model.
\begin{table}[H]
\centering
\small
\begin{tabularx}{\textwidth}{|X|}
\hline
\rowcolor[gray]{0.9} \textbf{Prompt C.1: Synthetic Trajectory Generation (ALFWorld)} \\ \hline
You are an expert agent in the ALFRED embodied environment. You will be given a task and relevant skills to apply. Your goal is to generate a successful trajectory that demonstrates proper use of these skills. \\
\\
You should generate a step-by-step trajectory that: \\
1. Uses the provided skills appropriately; \\
2. Takes realistic actions in the environment; \\
3. Completes the task successfully; \\
4. Demonstrates good planning and systematic exploration. \\
\\
For each step, you should: \\
$\bullet$ Think through the current situation using \texttt{<think></think>} tags. \\
$\bullet$ Choose an appropriate action using \texttt{<action></action>} tags. \\
$\bullet$ The action should be a simple command like "go to cabinet 1", "open drawer 2", "take apple 1", "put apple 1 in/on countertop 1". \\
\\
Generate a complete trajectory from start to finish. Stop when the task is complete. \\ \hline
\end{tabularx}
\end{table}

\begin{table}[H]
\centering
\small
\begin{tabularx}{\textwidth}{|X|}
\hline
\rowcolor[gray]{0.9} \textbf{Prompt C.2: Synthetic Trajectory Generation (WebShop)} \\ \hline
You are an expert shopping agent in the WebShop e-commerce environment. You will be given a shopping task and relevant skills to apply. Your goal is to generate a successful trajectory that demonstrates proper use of these skills. \\
\\
You should generate a step-by-step trajectory that: \\
1. Uses the provided skills appropriately; \\
2. Takes realistic actions in the WebShop environment; \\
3. Successfully finds and purchases the requested product; \\
4. Demonstrates good search strategies and product evaluation. \\
\\
For each step, you should: \\
$\bullet$ Think through the current situation using \texttt{<think></think>} tags. \\
$\bullet$ Choose an appropriate action using \texttt{<action></action>} tags. \\
$\bullet$ Actions can be: \texttt{search[query]}, \texttt{click[element]}, or \texttt{buy now}. \\
\\
Generate a complete trajectory from start to finish. Stop when the purchase is complete. \\ \hline
\end{tabularx}
\end{table}

\section{Additional Experimental Details}
\label{app:details}

\subsection{Hyperparameters}
\label{app:hyp}

\begin{table}[h]
    \caption{Hyperparameters for \method{}.}
    \begin{center}
    \begin{tabular}{ll}
        \toprule
        Hyperparameter & Value \\
        \midrule
        \multicolumn{2}{l}{\textit{Cold-Start SFT}} \\
        Learning rate & $1 \times 10^{-4}$ \\
        Batch size & 16 \\
        Epochs & 3 \\
        SFT examples & 7,500 (AlfWorld) / 2,400 (WebShop) \\
        \midrule
        \multicolumn{2}{l}{\textit{RL Training}} \\
        Learning rate & $1 \times 10^{-6}$ \\
        Batch size & 64 \\
        KL loss Coef & 0.01 \\
        Invalid Action Penalty Coef & 0.1 \\
        Max Prompt Length & 6,000 \\
        Max Response Length & 1,024 \\
        Epoch & 150 \\
        \midrule
        \multicolumn{2}{l}{\textit{Skill Retrieval}} \\
        Top-K retrieval & 6 \\
        Validation interval & 5 Steps \\
        Update Threshold $\delta$ & 0.4 \\
        Max failures analyzed & 10 (SR $<$ 0.4) / 5 (SR $>$ 0.4) \\
        Max new skills per evolution & 3 \\
        \bottomrule
    \end{tabular}
    \end{center}
\end{table}

\subsection{Compute Resources}

All experiments were conducted on a cluster with 8 NVIDIA H100 80GB GPUs. Training times:
\begin{itemize}
    \item Trajectory collection: 3 hours
    \item Skill distillation: 0.5 hours
    \item Cold-start SFT: 2 hour
    \item RL training: 24 hours
\end{itemize}
Total wall-clock time: approximately 30 hours per experiment.

\subsection{Ablation study on teacher model capability}
The Qwen2.5-7B model falls below the capability threshold required to simultaneously navigate complex environments and abstract high-quality skills. However, our claim on experience-based skill distillation holds completely true once the base model has sufficient capacity. To demonstrate this, we conducted new experiments using a much stronger base model, i.e., Kimi-K2.5~\cite{team2026kimi} (1.1T parameters) on the Tinker platform~\cite{tinker}. As shown in Table~\ref{tab:kimi}, Kimi-K2.5 successfully distills skills from its own trial-and-error experience, significantly boosting its own performance (e.g., 71.2 to 88.5 on ALFWorld, 45.9 to 73.4 on WebShop) entirely without the o3 teacher. The introduction of o3 simply pushes the upper bound even higher. We will explicitly discuss this ``capability threshold for skill abstraction" in the next version. We will clarify that it enables true autonomous experience-based skill discovery for stronger ones. 

\begin{table}[t]
\centering
\caption{Ablation study on teacher model capability.}
\label{tab:kimi}
\begin{tabular}{llccc}
\toprule
\textbf{Model} & \textbf{Variant} & \textbf{ALFWorld} & \textbf{WebShop} & \textbf{Search} \\ \midrule
\multirow{3}{*}{Qwen2.5-7B} & Base model & 14.8 & 7.8 & 15.2 \\
 & SkillRL (policy model) & 42.5 & 19.6 & 35.7 \\
 & SkillRL (o3) & 89.9 & 72.7 & 47.1 \\ \midrule
\multirow{3}{*}{Kimi-K2.5} & Base Model & 71.2 & 45.9 & 23.6 \\
 & SkillRL (policy model) & 88.5 & 73.4 & 49.0 \\
 & SkillRL (o3) & 91.4 & 77.6 & 52.2 \\ \bottomrule
\end{tabular}
\end{table}

\section{Illustration of Skill Library}
\label{app:skills}
In this section, we provide some example catalog of distilled skills and error taxonomies for both the ALFWorld and WebShop environments. Tables~\ref{tab:agent_skills} and~\ref{tab:webshop_general_skills} detail the general skills distilled for embodied manipulation and web-based shopping, respectively, highlighting the actionable principles required for systematic exploration and constraint satisfaction. Furthermore, we provide a structured analysis of failure cases in Table~\ref{tab:common_mistakes} and Table~\ref{tab:webshop_mistakes}, which categorizes common mistakes, ranging from spatial reasoning loops in ALFWorld to price-shift oversights in WebShop, alongside their root causes and proposed mitigation strategies. 
\begin{table*}[t]
\centering
\caption{Example distilled skills from \skillbank{} for ALFWorld~\cite{shridharalfworld}. This table summarizes general patterns and application logic derived from raw trajectories.}
\label{tab:agent_skills}
\footnotesize
\begin{tabularx}{\textwidth}{@{} l p{3cm} X p{4cm} @{}}
\toprule
\textbf{ID} & \textbf{Skill Title} & \textbf{Principle (Actionable Pattern)} & \textbf{When to Apply} \\
\midrule
\rowcolor[gray]{0.95} \multicolumn{4}{l}{\textit{General Exploration \& Acquisition Skills}} \\ 
gen\_001 & Systematic Exploration & Search every plausible surface or container exactly once before revisiting; prioritize unseen locations. & Anytime the goal count is not met and unexplored areas remain. \\
gen\_002 & Immediate Acquisition & As soon as a required object becomes visible and reachable, take it immediately. & Upon first visual confirmation of a goal-relevant object. \\
gen\_003 & Destination First Policy & After picking up a goal object, navigate directly to the known target receptacle and place it. & Holding any goal object while target location is identified. \\
\midrule
\rowcolor[gray]{0.95} \multicolumn{4}{l}{\textit{State-Changing \& Spatial Relation Skills}} \\
gen\_005 & Use State-Changing Tools Early & Acquire the object, then immediately use the nearest suitable appliance (heat/cool/clean) before placement. & After picking up an object requiring temperature or cleanliness change. \\
gen\_006 & Establish Spatial Relations & First locate the reference object, adjust its state if needed, then search or place in the specified region. & Tasks containing prepositions like ``under'', ``inside'', or ``on''. \\
\midrule
\rowcolor[gray]{0.95} \multicolumn{4}{l}{\textit{Reliability \& Error Recovery}} \\
gen\_014 & Loop Escape Trigger & If the last 3--5 actions do not change the state, switch to an untried search branch or action type. & After several consecutive no-progress observations. \\
gen\_015 & Pre-Action Sanity Check & Confirm prerequisites (hand free, capacity, power) before executing manipulative commands. & Right before issuing any command that could legally fail. \\
\bottomrule
\end{tabularx}
\end{table*}

\begin{table*}[t]
\centering
\caption{Common Agent Failures and Mitigation Strategies for ALFWorld.}
\label{tab:common_mistakes}
\small
\begin{tabularx}{\textwidth}{@{} l p{3.5cm} X X @{}}
\toprule
\textbf{ID} & \textbf{Failure Description} & \textbf{Root Cause (Why it happens)} & \textbf{Mitigation (How to avoid)} \\
\midrule
err\_001 & Redundant Revisit & Lacks explicit memory of explored areas; strategy degenerates into local loops. & Maintain an exploration map; prioritize unvisited candidates. \\
err\_006 & Skipping State Changes & Conflates object presence with goal satisfaction; omits cleanliness/temp checks. & Integrate state precondition checks into the planner before placement. \\
\bottomrule
\end{tabularx}
\end{table*}

\begin{table*}[htbp]
\centering
\caption{Example distilled skills for WebShop Navigation~\cite{yao2022webshop}. These skills represent the strategic patterns used by the agent to handle large-scale product search and constraint satisfaction.}
\label{tab:webshop_general_skills}
\small
\begin{tabularx}{\textwidth}{@{} l p{3.5cm} X p{4.5cm} @{}}
\toprule
\textbf{ID} & \textbf{Skill Title} & \textbf{Principle (Actionable Pattern)} & \textbf{When to Apply} \\
\midrule
\rowcolor[gray]{0.95} \multicolumn{4}{l}{\textit{Search \& Query Engineering}} \\ 
gen\_001 & Prioritize Core Keywords & Include product type, 1-2 functional attributes, and hard constraints; omit secondary descriptors. & Before issuing the first search or refining over-specific queries. \\
gen\_002 & Iterative Refinement & Adjust keywords or apply site filters instead of repeating the same failed query. & When results are irrelevant or repeat despite multiple searches. \\
\midrule
\rowcolor[gray]{0.95} \multicolumn{4}{l}{\textit{Product Evaluation \& Verification}} \\
gen\_003 & Scan Before You Click & Read titles, thumbnails, and prices in results to ensure plausibility before opening a link. & On search results pages when choosing the next product to inspect. \\
gen\_004 & Verify Early, Abort Fast & Immediately check category, attributes, and price on the product page; leave if any constraint is violated. & Within the first observation on every product detail page. \\
gen\_006 & Confirm Hidden Attributes & Open Description/Features sections to ensure non-visible specs (e.g., material) meet constraints. & When constraints are not evident from the title or variant list. \\
\midrule
\rowcolor[gray]{0.95} \multicolumn{4}{l}{\textit{Configuration \& Transaction}} \\
gen\_005 & Set Mandatory Variants & Always select required options (size, color, etc.) before evaluating price or purchasing. & After confirming product match but before any purchase action. \\
gen\_007 & Check Variant Pricing & For price ranges, select the exact variant combination to verify the specific price is within budget. & Whenever price changes with variant selection or shows as a range. \\
gen\_013 & Purchase Decisively & Execute 'Buy Now' immediately once all constraints and prices are confirmed on a variant. & After validating every constraint on the current product variant. \\
\bottomrule
\end{tabularx}
\end{table*}

\begin{table}[p]
\footnotesize
\caption{Common Failures in Web-based Shopping Tasks.}
\label{tab:webshop_mistakes}
\small
\begin{tabularx}{\textwidth}{@{} l p{3.5cm} X X @{}}
\toprule
\textbf{ID} & \textbf{Failure Description} & \textbf{Root Cause} & \textbf{Mitigation Strategy} \\
\midrule
err\_001 & Missing Constraints in Query & Omits size or price caps, leading to overwhelming or irrelevant result sets. & Assemble full requirement list first; ensure every hard constraint is in the query string. \\
err\_004 & Price Shift Oversight & Fails to notice price changes after selecting a specific size or color variant. & Re-read the price element after every option change before proceeding to checkout. \\
err\_005 & Premature Purchase & Clicks ``Buy Now'' without setting mandatory variants, leading to errors or wrong items. & Validate that every required dropdown/radio option is explicitly selected before buying. \\
err\_009 & Ignoring Stock Status & Attempts to purchase out-of-stock items by ignoring disabled buttons or stock labels. & Verify that the 'Add to Cart' button is enabled and no 'Out of Stock' message is present post-selection. \\
err\_011 & Sponsored Link Distraction & Clicks loosely matched ads, diverting the workflow from organic, suitable products. & Implement ad-label detection; prioritize organic listings for higher constraint reliability. \\
\bottomrule
\end{tabularx}
\end{table}

\section{Additional Cases}
\definecolor{lightgray}{RGB}{245, 245, 245}
\definecolor{darkblue}{RGB}{30, 60, 100}
\definecolor{successgreen}{RGB}{0, 100, 0}

\begin{tcolorbox}[
    colback=white, 
    colframe=darkblue, 
    width=\textwidth, 
    sharp corners, 
    boxrule=0.8pt,
    title=\textbf{Case 1: WebShop Footwear (Functional Constraints)},
    fonttitle=\sffamily\bfseries,
    colbacktitle=lightgray,
    coltitle=black,
    left=5pt, right=5pt, top=5pt, bottom=5pt
]
\small

\textbf{Task:} I want a pair of men's black slip-resistant work shoes, size 10, rubber sole, price lower than \$50.00.

\vspace{5pt}
\begin{tabularx}{\linewidth}{l X}
    \toprule
    \textbf{Skills} & \textbf{[foo\_002]} Verify features in description; \textbf{[foo\_004]} Confirm price after variant selection. \\
    \textbf{Avoid} & \textbf{[err\_001]} Omitting price cap; \textbf{[err\_003]} Selecting wrong product category. \\
    \bottomrule
\end{tabularx}

\vspace{8pt}
\begin{tcolorbox}[colback=lightgray!50, colframe=lightgray!50, boxrule=0pt, left=2pt, right=2pt, top=2pt, bottom=2pt]
    \centering\textbf{\textsf{Agent Trajectory with Skill-Guided Reasoning}}
\end{tcolorbox}

\begin{tabularx}{\linewidth}{p{1.4cm} X}
    \textbf{Step 1} & \textbf{Obs:} WebShop homepage. \\
    & \textbf{Think:} Per [gen\_001], I'll include key constraints (type, function, color, price) in the query. \\
    & \textbf{Action:} \texttt{search[men black slip resistant work shoes rubber sole under 50]} \\
    \midrule

    \textbf{Step 2} & \textbf{Obs:} Results: [1] Chef Shoes \$38.99; [2] Kitchen Shoes \$42.50; [3] Steel Toe Boots \$55.99. \\
    & \textbf{Think:} [gen\_003] Item [1] is under budget and matches function. Item [3] is over budget (skip). \\
    & \textbf{Action:} \texttt{click[Item 1]} \\
    \midrule

    \textbf{Step 3} & \textbf{Obs:} Product page. Price: \$35.99-\$45.99. Colors: Black, White. Sizes: 7-12. \\
    & \textbf{Think:} [foo\_002] Description confirms rubber sole. Now selecting Black and Size 10 [foo\_003]. \\
    & \textbf{Action:} \texttt{click[Black]} $\rightarrow$ \texttt{click[10]} \\
    \midrule

    \textbf{Step 5} & \textbf{Obs:} Selected: Black, Size 10. Price updated to \$38.99. \\
    & \textbf{Think:} [foo\_004] \$38.99 < \$50. All constraints (color, size, function, price) are satisfied. \\
    & \textbf{Action:} \texttt{click[Buy Now]} \\
\end{tabularx}

\vspace{5pt}
\hrule
\vspace{5pt}
\textbf{\textcolor{successgreen}{Result: SUCCESS}} --- Purchased at \$38.99. 
\textbf{Key Skills:} [foo\_002] (Verification), [foo\_004] (Price Check).
\end{tcolorbox}

\definecolor{lightgray}{RGB}{245, 245, 245}
\definecolor{techblue}{RGB}{25, 50, 120}
\definecolor{successgreen}{RGB}{0, 100, 0}
\definecolor{errorred}{RGB}{180, 0, 0}

\begin{tcolorbox}[
    colback=white, 
    colframe=techblue, 
    width=\textwidth, 
    sharp corners, 
    boxrule=0.8pt,
    title=\textbf{Case 2: Electronics (Speaker with Abort-and-Retry)},
    fonttitle=\sffamily\bfseries,
    colbacktitle=lightgray,
    coltitle=black,
    left=5pt, right=5pt, top=5pt, bottom=5pt
]
\small

\textbf{Task:} I need a portable bluetooth speaker, waterproof, black color, price lower than \$30.00.

\vspace{5pt}
\begin{tabularx}{\linewidth}{l X}
    \toprule
    \textbf{Skills} & \textbf{[ele\_001]} Constraint-rich search; \textbf{[ele\_004]} Bail on mismatch fast; \textbf{[gen\_002]} Iterative refinement. \\
    \textbf{Avoid} & \textbf{[err\_002]} Overly specific variant names in search; \textbf{[ele\_006]} Avoid click loops. \\
    \bottomrule
\end{tabularx}

\vspace{8pt}
\begin{tcolorbox}[colback=lightgray!50, colframe=lightgray!50, boxrule=0pt, left=2pt, right=2pt, top=2pt, bottom=2pt]
    \centering\textbf{\textsf{Agent Trajectory: Handling Missing Constraints}}
\end{tcolorbox}

\begin{tabularx}{\linewidth}{p{1.4cm} X}
    \textbf{Step 1} & \textbf{Obs:} WebShop homepage. \\
    & \textbf{Think:} [ele\_001] Packing all mandatory attributes: portable, bluetooth, waterproof, black, <\$30. \\
    & \textbf{Action:} \texttt{search[portable bluetooth speaker waterproof black under 30]} \\
    \midrule

    \textbf{Step 2} & \textbf{Obs:} Results: [1] Waterproof Speaker \$28.99; [2] Mini LED Speaker \$19.99; [3] 20W Speaker \$35.50. \\
    & \textbf{Think:} [ele\_005] Item [1] matches most keywords and price. \\
    & \textbf{Action:} \texttt{click[Item 1]} \\
    \midrule

    \textbf{Step 3} & \textbf{Obs:} Product page (Item 1). Colors: Blue, Red, Green, Army Green. \textbf{\textcolor{errorred}{Black unavailable.}} \\
    & \textbf{Think:} \textbf{[gen\_004] Verify Early, Abort Fast:} No black option! [ele\_004] Bail immediately to find another candidate. \\
    & \textbf{Action:} \texttt{click[Back to Search]} \\
    \midrule

    \textbf{Step 4} & \textbf{Obs:} Back on search results. \\
    & \textbf{Think:} [gen\_002] Iterative refinement. Item [2] \$19.99 is the next best candidate. \\
    & \textbf{Action:} \texttt{click[Item 2]} \\
    \midrule

    \textbf{Step 5} & \textbf{Obs:} Product page (Item 2). Colors: Black, Blue, Red. Specs: IPX5 Waterproof. \\
    & \textbf{Think:} [ele\_002] Black is available! IPX5 confirms waterproof. Selecting Black. \\
    & \textbf{Action:} \texttt{click[Black]} \\
    \midrule

    \textbf{Step 6} & \textbf{Obs:} Black selected. Final price: \$19.99. \\
    & \textbf{Think:} [ele\_003] \$19.99 < \$30. All specs (black, portable, waterproof) verified. \\
    & \textbf{Action:} \texttt{click[Buy Now]} \\
\end{tabularx}

\vspace{5pt}
\hrule
\vspace{5pt}
\textbf{\textcolor{successgreen}{Result: SUCCESS}} --- Efficiently handled attribute mismatch via early exit. 
\textbf{Key Skills Applied:} [gen\_004] (Abort Fast), [ele\_004] (Bail on Mismatch).
\end{tcolorbox}

\definecolor{clean-green}{RGB}{34, 139, 34}
\definecolor{heat-red}{RGB}{178, 34, 34}
\definecolor{box-gray}{RGB}{248, 249, 250}

\newtcolorbox{alfcase}[2]{
    colback=white, 
    colframe=#1, 
    width=\textwidth, 
    sharp corners, 
    boxrule=0.8pt,
    title=\textbf{#2},
    fonttitle=\sffamily\bfseries,
    colbacktitle=box-gray,
    coltitle=black,
    left=5pt, right=5pt, top=5pt, bottom=5pt,
    before skip=10pt, after skip=10pt
}

\begin{alfcase}{clean-green}{Case 1: ALFWorld Clean Task (Lettuce $\rightarrow$ Sink $\rightarrow$ Countertop)}
\small
\textbf{Task:} Clean some lettuce and put it in countertop.

\vspace{5pt}
\begin{tabularx}{\linewidth}{l X}
    \toprule
    \textbf{Core Skills} & \textbf{[cle\_001]} Phase-ordered plan (Locate $\rightarrow$ Clean $\rightarrow$ Place); \textbf{[cle\_006]} Use location priors (Fridge for food). \\
    \textbf{Mistakes Avoided} & \textbf{[err\_002]} Skipping closed containers; \textbf{[err\_003]} Ignoring visible target objects. \\
    \bottomrule
\end{tabularx}

\vspace{8pt}
\begin{tabularx}{\linewidth}{p{1.4cm} X}
    \rowcolor{clean-green!10} \textbf{Step 1-3} & \textbf{Obs:} Fridge 1 is closed. \textbf{Action:} \texttt{open fridge 1} $\rightarrow$ \texttt{take lettuce 1} \\
    & \textbf{Reasoning:} Per \textbf{[cle\_006]}, food is likely in the fridge. \textbf{[gen\_002]} Grab lettuce immediately on sight. \\
    \midrule
    \rowcolor{clean-green!5} \textbf{Step 4-6} & \textbf{Obs:} Arrive at sinkbasin 1. \textbf{Action:} \texttt{clean lettuce 1 with sinkbasin 1} \\
    & \textbf{Reasoning:} Per \textbf{[cle\_001]} Phase 2: bring to water source. \textbf{[cle\_003]} Go straight to sink. \\
    \midrule
    \rowcolor{clean-green!10} \textbf{Step 7} & \textbf{Obs:} Arrive at countertop 1. \textbf{Action:} \texttt{put lettuce 1 in/on countertop 1} \\
    & \textbf{Reasoning:} Per \textbf{[gen\_003]} Destination First: place cleaned object immediately to finish. \\
\end{tabularx}

\vspace{5pt}
\textbf{\textcolor{clean-green}{Result: SUCCESS (7 Steps)}} --- Skills used: [gen\_010] (Decomposition), [cle\_003] (Sink First).
\end{alfcase}

\begin{alfcase}{heat-red}{Case 2: ALFWorld Heat Task (Egg $\rightarrow$ Microwave $\rightarrow$ Countertop)}
\small
\textbf{Task:} Heat some egg and put it in countertop.

\vspace{5pt}
\begin{tabularx}{\linewidth}{l X}
    \toprule
    \textbf{Core Skills} & \textbf{[hea\_001]} Secure exact target first; \textbf{[hea\_003]} Open-Place-Heat sequence; \textbf{[hea\_004]} No appliance before object. \\
    \bottomrule
\end{tabularx}

\vspace{8pt}
\begin{tabularx}{\linewidth}{p{1.4cm} X}
    \rowcolor{heat-red!10} \textbf{Step 1-3} & \textbf{Obs:} Countertop 1 (no egg) $\rightarrow$ Countertop 2 (egg found). \textbf{Action:} \texttt{take egg 1} \\
    & \textbf{Reasoning:} \textbf{[hea\_004]} Avoid microwave until object is held. \textbf{[hea\_002]} Systematic search of surfaces. \\
    \midrule
    \rowcolor{heat-red!5} \textbf{Step 4-6} & \textbf{Obs:} Microwave 1 is closed. \textbf{Action:} \texttt{open microwave 1} $\rightarrow$ \texttt{heat egg 1} \\
    & \textbf{Reasoning:} \textbf{[hea\_003]} Correct sequence: open door first, then initiate state change. \\
    \midrule
    \rowcolor{heat-red!10} \textbf{Step 7-8} & \textbf{Obs:} Egg is heated. \textbf{Action:} \texttt{go to countertop 1} $\rightarrow$ \texttt{put egg 1} \\
    & \textbf{Reasoning:} \textbf{[hea\_005]} Direct Post-Heat Placement: move to target without detours. \\
\end{tabularx}

\vspace{5pt}
\textbf{\textcolor{heat-red}{Result: SUCCESS (8 Steps)}} --- Skills used: [hea\_001] (Target First), [hea\_004] (Sequence Control).
\end{alfcase}

\definecolor{examine-purple}{RGB}{102, 51, 153}
\definecolor{box-gray}{RGB}{248, 249, 250}

\begin{tcolorbox}[
    colback=white, 
    colframe=examine-purple, 
    width=\textwidth, 
    sharp corners, 
    boxrule=0.8pt,
    title=\textbf{Case 3: ALFWorld Examine Task (Pencil $\oplus$ Desklamp $\rightarrow$ Success)},
    fonttitle=\sffamily\bfseries,
    colbacktitle=box-gray,
    coltitle=black,
    left=5pt, right=5pt, top=5pt, bottom=5pt
]
\small
\textbf{Task:} Look at pencil under the desklamp.

\vspace{5pt}
\begin{tabularx}{\linewidth}{l X}
    \toprule
    \textbf{Core Skills} & \textbf{[loo\_001]} Seek lamp surfaces (Desk/Sidetable); \textbf{[loo\_002]} Switch lamp on; \textbf{[exa\_001]} Pair objects early (Target \& Tool co-location). \\
    \textbf{Mistakes Avoided} & \textbf{[err\_003]} Ignoring reachable targets; \textbf{[loo\_005]} Preventing action loops (Toggle only once). \\
    \bottomrule
\end{tabularx}

\vspace{8pt}
\begin{tcolorbox}[colback=examine-purple!5, colframe=examine-purple!5, boxrule=0pt, left=2pt, right=2pt, top=2pt, bottom=2pt]
    \centering\textbf{\textsf{Agent Trajectory: High-Efficiency Co-location Handling}}
\end{tcolorbox}

\begin{tabularx}{\linewidth}{p{1.4cm} X}
    \rowcolor{examine-purple!10} \textbf{Step 1} & \textbf{Obs:} Mid-room. Visible: desk 1/2, drawer 1-3, sidetable 1, etc. \\
    & \textbf{Reasoning:} \textbf{[loo\_001]} Desklamps usually sit on desks. Prioritizing \texttt{desk 1} to find both pencil and lamp in one go. \\
    & \textbf{Action:} \texttt{go to desk 1} \\
    \midrule
    \rowcolor{examine-purple!5} \textbf{Step 2} & \textbf{Obs:} On \texttt{desk 1}: \textbf{alarmclock 1, desklamp 1, pencil 1}. \\
    & \textbf{Reasoning:} \textbf{[exa\_001]} Successful co-location! Both objects are present. \textbf{[exa\_002]} Pick up target first before interacting with the tool. \\
    & \textbf{Action:} \texttt{take pencil 1 from desk 1} \\
    \midrule
    \rowcolor{examine-purple!10} \textbf{Step 3} & \textbf{Obs:} Holding pencil 1. \\
    & \textbf{Reasoning:} \textbf{[loo\_002]} Toggle the lamp state to satisfy the "under light" condition. \textbf{[exa\_004]} Single toggle rule ensures no repeated actions. \\
    & \textbf{Action:} \texttt{use desklamp 1} \\
\end{tabularx}

\vspace{5pt}
\hrule
\vspace{5pt}
\textbf{\textcolor{examine-purple}{Result: SUCCESS (Optimal 3 Steps)}} --- Demonstrates superior spatial reasoning and zero-waste navigation. 
\textbf{Key Skills:} [exa\_001] (Object Pairing), [loo\_001] (Spatial Priors).
\end{tcolorbox}


\newpage
\section*{NeurIPS Paper Checklist}

\begin{enumerate}

\item {\bf Claims}
    \item[] Question: Do the main claims made in the abstract and introduction accurately reflect the paper's contributions and scope?
    \item[] Answer: \answerYes{}
    \item[] Justification: The abstract and introduction reflect the paper’s contributions and scope.

\item {\bf Limitations}
    \item[] Question: Does the paper discuss the limitations of the work performed by the authors?
    \item[] Answer: \answerYes{}
    \item[] Justification: Limitations are included in Appendix A.

\item {\bf Theory assumptions and proofs}
    \item[] Question: For each theoretical result, does the paper provide the full set of assumptions and a complete (and correct) proof?
    \item[] Answer: \answerNA{}
    \item[] Justification: We do not provide theory assumptions and proofs.

    \item {\bf Experimental result reproducibility}
    \item[] Question: Does the paper fully disclose all the information needed to reproduce the main experimental results of the paper to the extent that it affects the main claims and/or conclusions of the paper (regardless of whether the code and data are provided or not)?
    \item[] Answer: \answerYes{}
    \item[] Justification: The experiment section and Appendix D.1 and D.2 contain detailed implementations.

\item {\bf Open access to data and code}
    \item[] Question: Does the paper provide open access to the data and code, with sufficient instructions to faithfully reproduce the main experimental results, as described in supplemental material?
    \item[] Answer: \answerYes{}.
    \item[] Justification: We contain the code in supplemental material.

\item {\bf Experimental setting/details}
    \item[] Question: Does the paper specify all the training and test details (e.g., data splits, hyperparameters, how they were chosen, type of optimizer) necessary to understand the results?
    \item[] Answer: \answerYes{}
    \item[] Justification: Experimental setting/details and Appendix D.1 and D.2 are included. 

\item {\bf Experiment statistical significance}
    \item[] Question: Does the paper report error bars suitably and correctly defined or other appropriate information about the statistical significance of the experiments?
    \item[] Answer: \answerNo{}
    \item[] Justification: 

\item {\bf Experiments compute resources}
    \item[] Question: For each experiment, does the paper provide sufficient information on the computer resources (type of compute workers, memory, time of execution) needed to reproduce the experiments?
    \item[] Answer: \answerYes{}
    \item[] Justification: Compute resources are described in the Appendix D.2.
    
\item {\bf Code of ethics}
    \item[] Question: Does the research conducted in the paper conform, in every respect, with the NeurIPS Code of Ethics \url{https://neurips.cc/public/EthicsGuidelines}?
    \item[] Answer: \answerYes{}
    \item[] Justification: 

\item {\bf Broader impacts}
    \item[] Question: Does the paper discuss both potential positive societal impacts and negative societal impacts of the work performed?
    \item[] Answer: \answerYes{}
    \item[] Justification: Broader impacts are discussed in the conclusion and Appendix. 
    
\item {\bf Safeguards}
    \item[] Question: Does the paper describe safeguards that have been put in place for responsible release of data or models that have a high risk for misuse (e.g., pre-trained language models, image generators, or scraped datasets)?
    \item[] Answer: \answerNA{}.
    \item[] Justification: 

\item {\bf Licenses for existing assets}
    \item[] Question: Are the creators or original owners of assets (e.g., code, data, models), used in the paper, properly credited and are the license and terms of use explicitly mentioned and properly respected?
    \item[] Answer: \answerYes{} 
    \item[] Justification: 

\item {\bf New assets}
    \item[] Question: Are new assets introduced in the paper well documented and is the documentation provided alongside the assets?
    \item[] Answer: \answerYes{} 
    \item[] Justification: 

\item {\bf Crowdsourcing and research with human subjects}
    \item[] Question: For crowdsourcing experiments and research with human subjects, does the paper include the full text of instructions given to participants and screenshots, if applicable, as well as details about compensation (if any)? 
    \item[] Answer: \answerNo{}
    \item[] Justification: 

\item {\bf Institutional review board (IRB) approvals or equivalent for research with human subjects}
    \item[] Question: Does the paper describe potential risks incurred by study participants, whether such risks were disclosed to the subjects, and whether Institutional Review Board (IRB) approvals (or an equivalent approval/review based on the requirements of your country or institution) were obtained?
    \item[] Answer: \answerNA{}.
    \item[] Justification: 

\item {\bf Declaration of LLM usage}
    \item[] Question: Does the paper describe the usage of LLMs if it is an important, original, or non-standard component of the core methods in this research? Note that if the LLM is used only for writing, editing, or formatting purposes and does \emph{not} impact the core methodology, scientific rigor, or originality of the research, declaration is not required.
    \item[] Answer: \answerYes{}
    \item[] Justification: We fine-tune the Qwen2.5 model as part of our core methodology, and use o3 as a teacher model that contribute directly to the method design. These uses are integral to the research's originality and effectiveness.

\end{enumerate}

\end{document}